\documentclass[10pt,journal,compsoc]{IEEEtran}
\usepackage{times}
\usepackage{graphicx}
\usepackage{epstopdf}
\usepackage{epsfig}
\usepackage{amsmath}
\usepackage{amssymb}
\usepackage{algorithmic}
\usepackage{algorithm}
\usepackage{multirow}
\usepackage{textcomp}
\usepackage{xspace}
\usepackage{color}
\usepackage[tight,scriptsize,sf,SF]{subfigure}
\usepackage[it,small]{caption}
\usepackage{array}
\usepackage{enumitem}
\usepackage{hyperref}

\ifCLASSOPTIONcompsoc
  \usepackage[nocompress]{cite}
\else
  \usepackage{cite}
\fi
\ifCLASSINFOpdf
\else
\fi
\def \D {\mathbb{D}}
\def \A {\mathbb{A}}

\def \z {\mathbf{z}}
\def \w {\mathbf{w}}

\def \0 {\mathbf{0}}
\def \1 {\mathbf{1}}

\makeatletter
\DeclareRobustCommand\onedot{\futurelet\@let@token\@onedot}
\def\@onedot{\ifx\@let@token.\else.\null\fi\xspace}

\def\eg{\emph{e.g}\onedot} 
\def\ie{\emph{i.e}\onedot} 
 
\def\etc{\emph{etc}\onedot} 
 
\def\etal{\emph{et al}\onedot} \def\st{s.t\onedot}

\newcommand{\header}[1]{\smallskip\noindent\textbf{#1}}

\newcolumntype{R}[1]{>{\raggedright\arraybackslash\hspace{0pt}}m{#1}}

\begin{document}
%
\title{Watch-n-Patch: Unsupervised Learning of Actions and Relations}
%
%
%
%

\author{Chenxia Wu, 
        Jiemi Zhang, 
        Ozan Sener, 
        Bart Selman, 
        Silvio Savarese, 
        and~Ashutosh Saxena 
\IEEEcompsocitemizethanks{\IEEEcompsocthanksitem Wu, Sener and Selman are with the Department
of Computer Science, Cornell University, Ithaca, NY 14853.  
E-mail: {chenxiawu,ozan,selman}@cs.cornell.edu \protect\\
\IEEEcompsocthanksitem Zhang is with Didi Chuxing, China. Email: jmzhang10@gmail.com\protect\\
\IEEEcompsocthanksitem Savarese is with the Department of Computer Science, Stanford University, CA 94305. Email: ssilvio@cs.stanford.edu\protect\\

\IEEEcompsocthanksitem Saxena is with Brain of Things Inc., Redwood City, CA 94062. 
Email: asaxena@cs.stanford.edu \protect\\}
 }

\IEEEtitleabstractindextext{%

\begin{abstract}
There is a large variation in the activities that humans perform in their everyday lives. We consider modeling these composite human activities which comprises multiple basic level actions in a completely unsupervised setting. Our model learns high-level co-occurrence and temporal relations between the actions. We consider the video as a sequence of short-term action clips, which contains human-words and object-words. An activity is about a set of action-topics and object-topics indicating which actions are present and which objects are interacting with. We then propose a new probabilistic model relating the words and the topics. It allows us to model long-range action relations that commonly exist in the composite activities, which is challenging in previous works.

We apply our model to the unsupervised action segmentation and clustering, and to a novel application that detects forgotten actions, which we call action patching.  For evaluation, we contribute a new challenging RGB-D activity video dataset recorded by the new Kinect v2, which contains several human daily activities as compositions of multiple actions interacting with different objects. Moreover, we develop a robotic system that watches people and reminds people by applying our action patching algorithm. Our robotic setup can be easily deployed on any assistive robot.\thanks{Parts of this work have been published in~\cite{Wu_2015_CVPR,Wu_2016_ICRA} as the conference version.}

\end{abstract}

}

\maketitle

\IEEEdisplaynontitleabstractindextext

%
\IEEEpeerreviewmaketitle

  \abovedisplayskip 3.0pt plus2pt minus2pt%
 \belowdisplayskip \abovedisplayskip
\renewcommand{\baselinestretch}{0.98}

\newenvironment{packed_enum}{
\begin{enumerate}
  \setlength{\itemsep}{0pt}
  \setlength{\parskip}{0pt}
  \setlength{\parsep}{0pt}
}
{\end{enumerate}}

\newenvironment{packed_item}{
\begin{itemize}
  \setlength{\itemsep}{0pt}
  \setlength{\parskip}{0pt}
  \setlength{\parsep}{0pt}
}{\end{itemize}}

\newlength\savedwidth
\newcommand\whline[1]{\noalign{\global\savedwidth\arrayrulewidth
                               \global\arrayrulewidth #1} %
                      \hline
                      \noalign{\global\arrayrulewidth\savedwidth}}
\renewcommand\multirowsetup{\centering}

\newlength{\sectionReduceTop}
\newlength{\sectionReduceBot}
\newlength{\subsectionReduceTop}
\newlength{\subsectionReduceBot}
\newlength{\abstractReduceTop}
\newlength{\abstractReduceBot}
\newlength{\captionReduceTop}
\newlength{\captionReduceBot}
\newlength{\subsubsectionReduceTop}
\newlength{\subsubsectionReduceBot}
\newlength{\headerReduceTop}
\newlength{\figureReduceBot}

\newlength{\horSkip}
\newlength{\verSkip}

\newlength{\equationReduceTop}

\newlength{\figureHeight}
\setlength{\figureHeight}{1.7in}

\setlength{\horSkip}{-.09in}
\setlength{\verSkip}{-.1in}

\setlength{\figureReduceBot}{-0.15in}
\setlength{\headerReduceTop}{0in}
\setlength{\subsectionReduceTop}{-0.02in}
\setlength{\subsectionReduceBot}{-0.02in}
\setlength{\sectionReduceTop}{-0.02in}
\setlength{\sectionReduceBot}{-0.01in}
\setlength{\subsubsectionReduceTop}{-0.06in}
\setlength{\subsubsectionReduceBot}{-0.05in}
\setlength{\abstractReduceTop}{-0.05in}
\setlength{\abstractReduceBot}{-0.10in}

\setlength{\equationReduceTop}{-0.1in}

\setlength{\captionReduceTop}{-0.06in}
\setlength{\captionReduceBot}{-0.07in}

\section{Introduction}\label{sec:intro}

\begin{figure}[t]
  \begin{center}
  \includegraphics[width=0.9\linewidth]{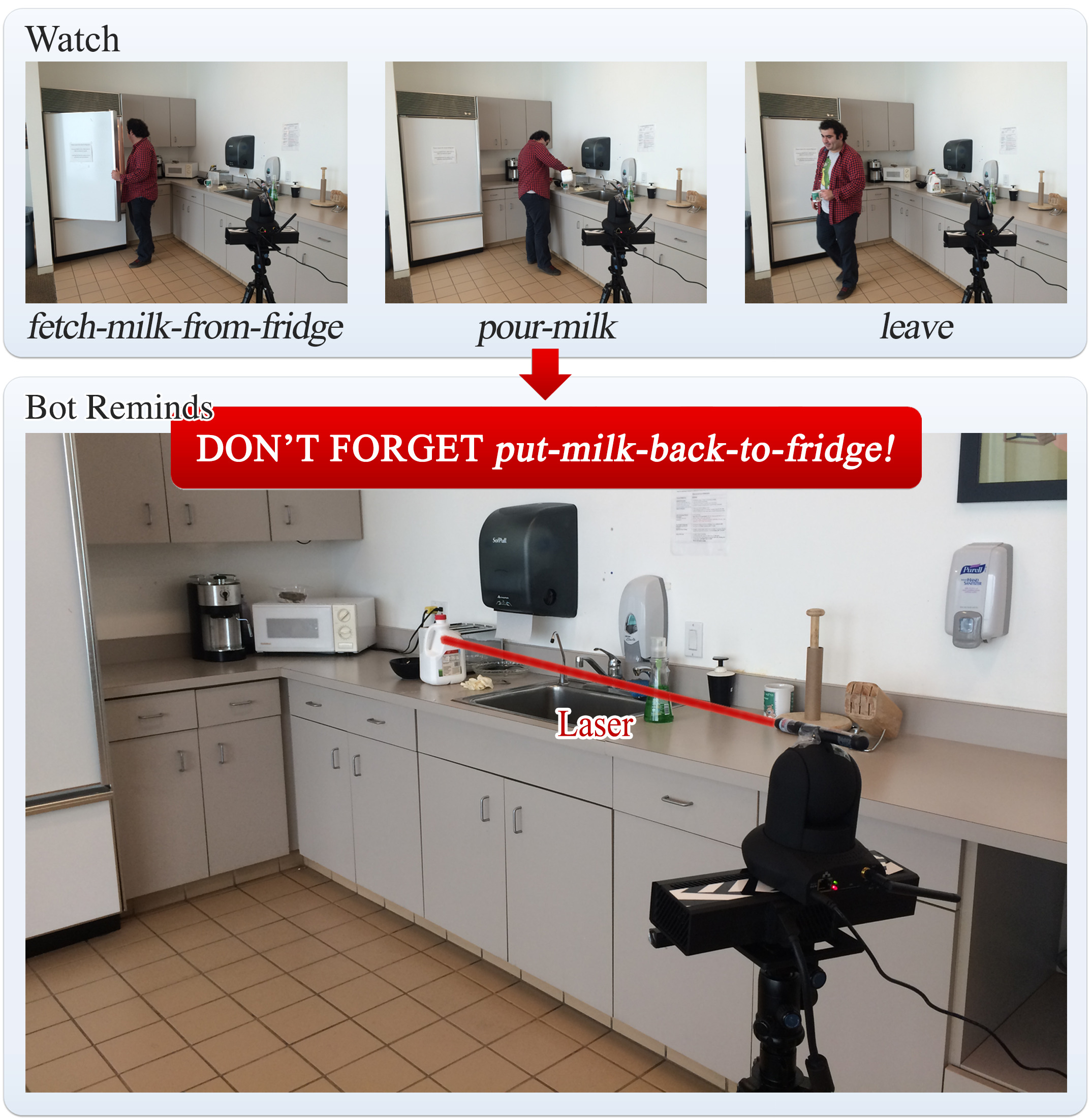}
   \caption{Our Watch-Bot understands what human is currently doing by automatically segmenting the composite activity into basic level actions. We propose a completely unsupervised approach to modeling the human skeleton and object features to the actions, as well as the pairwise action co-occurrence and temporal relations. Using the learned model, our robot detects humans' forgotten actions and reminds them by pointing out the related object using the laser spot.} 
   
 \label{fig:tr}
 \vspace{-0.1in}
 \end{center}
\end{figure}
 
The average adult forgets three key facts, chores or events every day~\cite{forget}. Hence it is important for a vision system to be able to detect not only what a human is currently doing but also what he forgot to do. For example in Fig.~\ref{fig:tr}, someone fetches milk from the fridge, pours the milk to the cup, takes the cup and leaves without putting back the milk, then the milk would go bad. In this paper, we focus on modeling these \emph{composite human activities} then detecting the \emph{forgotten actions} for a robot, which learns from a completely unlabeled set of RGB-D videos.

A human activity is composite, \ie, it is composed of several basic level actions. For example, a composite activity \emph{warming milk} contains a sequence of actions:  \emph{fetch-milk-from-fridge}, \emph{microwave-milk}, \emph{put-milk-back-to-fridge}, \emph{fetch-milk-from-microwave}, and \emph{leave}.  
Modeling this poses several challenges. First, some actions often co-occur in a composite activity but some may not. Second, co-occurring actions have variations in temporal orderings, \eg, people can first \emph{put-milk-back-to-fridge} then \emph{microwave-milk} instead of the inverse order in the above example, as its ordering is more relevant to the action \emph{fetch-milk-from-fridge}. Moreover, these ordering relations could exist in both short-range and long-range, \eg, \emph{pouring} is followed by \emph{drink} while sometimes \emph{fetch-book} is related to \emph{put-back-book} with a long \emph{read} between them. Third, the objects the human interacting with are also important to modeling the actions and their relations, as same actions often have common objects in interaction.

The challenge that we undertake in this paper is: Can an algorithm learn about the aforementioned relations in the composite activities when just given a completely \emph{unlabeled} set of \mbox{RGB-D} videos?


Most previous works focus on action detection in a supervised learning setting. In the training, they are given fully labeled actions in videos~\cite{Liu_2011_CVPR,Sadanand_2012_CVPR,Schiele_2012_CVPR},  or weakly supervised action labels~\cite{Duchenne_2009_ICCV,Bojanowski_2014_ECCV}, or locations of human/their interacting objects~\cite{Laptev_2007_ICCV,Tian_2013_CVPR,Ni_2014_CVPR}. Among them, the temporal structure of actions is often discovered by Markov models such as Hidden Markov Model (HMM)~\cite{Tang_2012_CVPR} and semi-Markov~\cite{Hoai_2011_CVPR, Shi_2011_IJCV}, or by linear dynamical systems~\cite{Bhattacharya_2014_CVPR}, or by hierarchical grammars~\cite{Pirsiavash_2014_CVPR,Vo_2014_CVPR,Kuehne_2014_CVPR,Wang_2014_CVPR,Assari_2014_CVPR}, or by other spatio-temporal representations~\cite{Ke_2007_ECCV,Niebles_2010_ECCV,Klaser_2010_ECCV,Koppula_2013_ICML}. Object-in-use contextual information has also been commonly used for recognizing actions~\cite{Koppula_2013_RSS,Koppula_2013_ICML,Ni_2014_CVPR,Wang_2014_CVPR}. Besides relying on the manually labeling, most of these works are based on RGB features and only model the short-range relations between actions (see Section~\ref{sec:re} for details). 


Unlike these approaches, we consider a completely unsupervised setting. The novelty of our approach is the ability to model the long-range action relations in the temporal sequence,
by considering pairwise action co-occurrence and temporal relations, \eg, \emph{put-milk-back-to-fridge} often co-occurs with and temporally after (but not necessarily follows) \emph{fetch-milk-from-fridge}. We also use the more informative human skeleton features and \mbox{RGB-D} object features, which have shown higher performance
over RGB only features for action recognition~\cite{Koppula_2013_RSS,Wu_2014_CVPR,Lin_2014_CVPR}.

\begin{figure}[t]
  \begin{center}
  \includegraphics[width=1\linewidth]{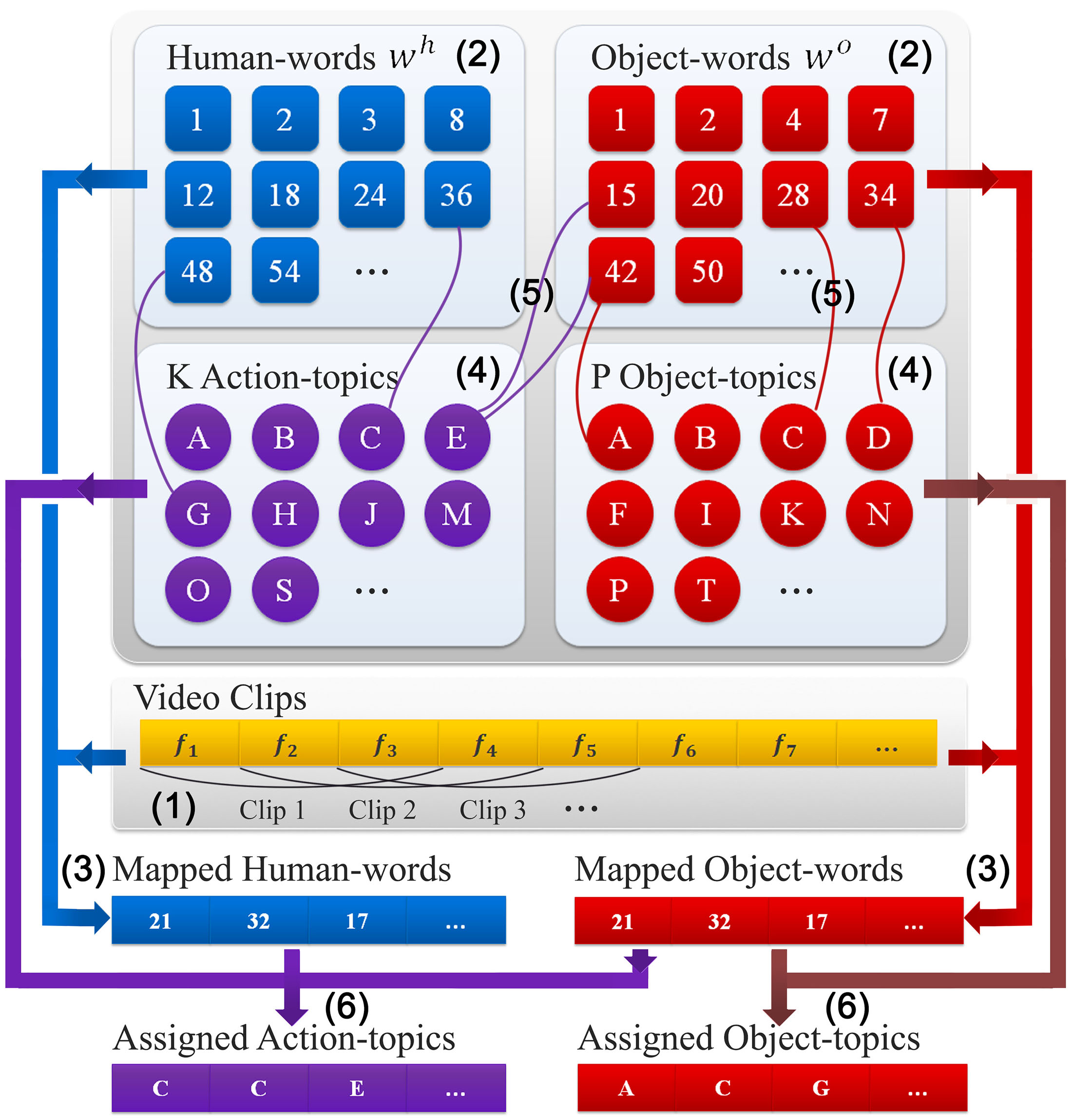}

 \caption{Video representation. (1) A video frames ($f_i$) is first decomposed into a sequence of overlapping fixed-length temporal clips. (2) The human-skeleton-trajectories/interactive-object-trajectories are extracted from each clip, and we cluster them to form the human-dictionary/object-dictionary. (3) Then the video is represented as a sequence of human-word and object-word indices by mapping its human-skeleton-trajectories/interactive-object-trajectories to the nearest human-words/object-words in the dictionary. (4) An activity video is about a set of action-topics/object-topics indicating which actions are present and which types of objects are interacting with. (5) We learn the mapping of action-words/object-words to the action-topics/object-topics, as well as the co-occurrence and the temporal relations between the topics. (6) We assign the topics to clips using the learned model.} \label{fig:vrep}
 \end{center}
 \vspace{-0.1in}
\end{figure}

In order to capture the rich structure in the composite activity, we draw strong parallels with the work done on document modeling from natural language (\eg, \cite{Blei_2003_LDA}) and proposed a Casual Topic Model (CaTM). We consider an activity video as a document,  which consists of a sequence of short-term action clips containing human-skeleton-trajectories as \emph{human-words} and interacting-object-trajectories as \emph{object-words}. An activity is about a set of \emph{action-topics} indicating which actions are present in the video, such as \emph{fetch-milk-from-fridge} in the \emph{warming milk} activity, and a set of \emph{object-topics} indicating which object types are interacting. We draw human-words from the action-topics, and object-words from both action-topics and object-topics\footnote{Here we consider the same object type like book can be variant in appearance in different actions such as close book in the fetch-book action and open book in the reading action.}. Then we model the following (see Fig.~\ref{fig:vrep}):

\begin{packed_item}

\item \emph{Action co-occurrence.}  Some actions often co-occur in the same activity and may have the same objects. We model the co-occurrence by adding correlated topic priors to the occurrence of action-topics and object-topics, \eg, action-topics \emph{fetch-book} and \emph{put-back-book} has strong correlations and are also strongly correlated to object-topic \emph{book}.
\item \emph{Action temporal relations.}   Some actions often causally follow each other, and actions change over time during the activity execution. We model the relative time distributions between every action-topic pair to capture the temporal relations.


\end{packed_item}

We first show that our model is able to learn meaningful representations from the unlabeled composite activity videos. We use the model to temporally segment videos to action segments by assigning action-topics. We show that these action-topics are promising to be semantically meaningful by mapping them to ground-truth action classes and evaluating the labeling performance.

We then show that our model can be used to detect forgotten actions in the composite activity, a new application that we call \emph{action patching}. We enable a robot, which we call \emph{Watch-Bot}, to detect humans' forgotten actions as well as to localize the related object in the scene. The setup of the robot can be easily deployed on any assistive robot and applied to different areas such as industry, medical work and home use. We evaluate the action patching accuracy to show that the learned co-occurrence and temporal relations are very helpful to inferring the forgotten actions. We also show that our Watch-Bot is able to remind humans of forgotten actions in the real-world robotic experiments.

We also provide a new challenging \mbox{RGB-D} activity video dataset~\footnote{The dataset and tools are released at http://watchnpatch.cs.cornell.edu.} recorded by the new Kinect v2  (see examples in Fig.~\ref{fig:actions}), in which the human skeletons are also recorded. It contains $458$ videos of human daily activities as compositions of multiple actions interacting with different objects, in which people forget actions in $222$ videos. They are performed by different subjects in different environments with complex backgrounds.  In robotic experiments, we show that our Watch-Bot is able to remind humans of forgotten actions in the real-world experiments.

In summary, the main contributions of this work are:
\begin{packed_item}
\item Our model is completely unsupervised thus being more useful and scalable.
\item Our model considers both the short-range and the long-range action relations, showing the effectiveness in the action segmentation and clustering.
\item We show a new application by enabling a robot to remind humans of forgotten actions in the real scenes.
\item We provide a new challenging \mbox{RGB-D} activity dataset recorded by the new Kinect v2, which contains videos of multiple actions interacting with different objects.
\end{packed_item}

The paper is organized as follows. Section~\ref{sec:re} introduces the related works. Section~\ref{sec:ov} outlines our approach to modeling the composite activity. We present the visual features of the activity video clip in Section~\ref{sec:fea}. Section~\ref{sec:apc} gives the detailed description of our learning model as well as its learning and inference. Section~\ref{sec:ap} introduces our watch-bot system to reminding of forgotten actions using our learned model. We give an extensive evaluation and discussion in the experiments in Section~\ref{sec:exp}. Sections~\ref{sec:con} concludes the paper. 

\if
We focus on simultaneously segmenting and recognizing multiple human actions from a video in an unsupervised way. This is more natural and critical than traditional action recognition which focuses on classification of pre-segmented single action clips. These works ignore the semantic and temporal relations between actions. However, in a real-world action sequence, some high-level actions often have strong correlations and consistent temporal ordering. For example, \emph{turn-off monitor} often comes with \emph{turn-on monitor}. On the other hand, some actions are independent or weakly correlated and vary in duration and temporal ordering such as \emph{drinking} and \emph{reading}.

In this paper, we develop an unsupervised probabilistic model to simultaneously segmenting each video into multiple actions and clustering these human actions. The model not only estimates the probabilities of the occurrence of actions in a video by modeling the correlations between actions, but also estimates the ordering of occurring actions by modeling their relative time. As a result, the semantic and temporal relations between actions in the whole video are fully discovered and represented. Using the learned model, we can further segment a novel input video and assign its segments to action clusters. Moreover, based on the learned action correlations and ordering, we can even infer forgotten actions, called \emph{action patching}. This is a new significant application of action recognition, since people would never worry about forgetting things in daily life.

As there is few datasets containing continuous high-level action sequences and action forgotten cases, we collect a new challenging \mbox{RGB-D} action dataset recorded by new Kinect v2. Each video in the dataset contains a long action sequence interacted with different objects (see examples in Fig.~\ref{fig:actions}). 
Some actions occur together often while some are not necessary in the same video. Some actions are in fix ordering while some occur in random order. Moreover, to evaluate the action patching performance, we also record videos with action forgotten by people naturally (see examples in Table~ and Fig~). The extensive experiments on this dataset show that our approach outperforms the state-of-the-art on both action segmentation and patching.

In summary, the main contributions of this work are:

\begin{itemize}
\item We propose a novel unsupervised probabilistic model to simultaneously segmenting each video into multiple actions and clustering them in consideration of both semantic and temporal relations.
\item We use our model to achieve a new significant application that reminds people of forgotten things, which only depends on Kinect sensor without any human supervision.
\item We provide a new challenging \mbox{RGB-D} action dataset recorded by the new Kinect v2, which contains videos of long action sequences interacted with different objects.
\end{itemize}
\fi

\section{Related Work}\label{sec:re}
\header{Action Recognition.} Our work is related to the works on action recognition in computer vision. There is a large number of works on action recognition, which can be referred in recent surveys~\cite{Aggarwal_2011_SURVEY}. In this section, we cover the most related approaches. Most previous works on action recognition are supervised~\cite{Laptev_2007_ICCV,Duchenne_2009_ICCV,Niebles_2010_ECCV,Liu_2011_CVPR,Sadanand_2012_CVPR,Tian_2013_CVPR,Bojanowski_2014_ECCV,Mathe_2014_PAMI}. Among them, the most popular are linear-chain models such as hidden markov model (HMM)~\cite{Tang_2012_CVPR}, semi-Markov~\cite{Hoai_2011_CVPR, Shi_2011_IJCV} and the linear dynamic system~\cite{Bhattacharya_2014_CVPR}. They focus on modeling the local transitions (between frames, temporal segment, or sub-actions) in the activities. More complex hierarchical relations~\cite{Pirsiavash_2014_CVPR,Vo_2014_CVPR,Kuehne_2014_CVPR,Wang_2014_CVPR} or graph relations~\cite{Assari_2014_CVPR,Souza_2015_CVPR} are considered in modeling actions in the complex activity.
There are also some works focusing on detecting local action patches, primitives, trajectories or spatio-temporal features~\cite{Jain_2013_CVPR,Yang_2013_TPAMI,Narayan_2014_CVPR,Ma_2015_CVPR} without considering the high-level action relations. There also exist some unsupervised approaches on action recognition. Yang \etal~\cite{Yang_2013_TPAMI} develop a meaningful representation by discovering local motion primitives in an unsupervised way, then a HMM is learned over these primitives. Jones \etal~\cite{Jones_2014_CVPR} propose an unsupervised dual assignment clustering on the dataset recorded from two views.  

Although these approaches have performed well in different areas, most of them rely on local relations between adjacent clips or actions that ignore the long-term action relations and use RGB visual features.  Unlike these approaches, we use the richer human skeleton and \mbox{RGB-D} features rather than the RGB action features~\cite{Wang_2011_CVPR,Kantorov_2014_CVPR}. We model the pairwise action co-occurrence and temporal relations in the whole video, thus relations are considered globally and completely with the uncertainty.  We also use the learned relations to infer the forgotten actions without any manual annotations.

\header{RGB-D and Human Skeleton Features.} Action recognition using human skeletons and \mbox{RGB-D} camera have shown the advantages over RGB videos in many works. Skeleton-based approach focus on proposing good skeletal representations~\cite{Schiele_2012_CVPR, Sung_2012_ICRA,Vemulapalli_2014_CVPR,Wu_2014_CVPR,Lin_2014_CVPR}. Besides of the human skeletons, we also detect the human interactive objects in an unsupervised way to provide more discriminate features. Object-in-use contextual information has been commonly used for recognizing actions~\cite{Koppula_2013_RSS,Koppula_2013_ICML,Ni_2014_CVPR,Wang_2014_CVPR}.  Moreover, Hu\etal~\cite{Hu_2015_CVPR} propose a joint learning model to simultaneously
learn heterogenous features from RGB-D activity videos. Most of them focus on designing or learning good action features. They lost the high-level action relations which can be captured in our model.

\header{Bayesian Models.} Our work is also related to the Bayesian models. LDA~\cite{Blei_2003_LDA} was the first hierarchical Bayesian topic model and widely used in different applications. The correlated topic models~\cite{Blei_2007_CTM,Kim_2011_NIPS} add the priors over topics to capture topic correlations. A topic model 
over absolute timestamps of words is proposed in~\cite{Wang_2006_KDD} and has been applied to action recognition~\cite{Tanveer_2009_BMVC}. However, the independence assumption of different topics would lead to non smooth temporal segmentations. Recently, a multi-feature max-margin hierarchical Bayesian model~\cite{Yang_2015_CVPR} is proposed to jointly learn a high-level representation by combining a hierarchical generative model and discriminative maxmargin classifiers in a unified Bayesian framework. Differently, our model considers both correlations and the relative time distributions between topics rather than the absolute time, which captures richer information of action structures in the complex human activity.

\header{Perception of Human Activities for Robotics.} Our work is also related to the works on recognizing human actions for robotics~\cite{Losch_2007_RO,Koppula_2013_IJRR,Chen_2014_ICRA}. Yang~\etal.~\cite{Yang_2015_AAAI} presented a system that learns manipulation action plans for robot from unconstrained youtube videos. Hu~\etal.~\cite{Hu_2014_RSS} proposed an activity recognition system trained from soft labeled data for the assistant robot. Chrungoo~\etal.~ \cite{Chrungoo_2014_SR} introduced a human-like stylized gestures for better human-robot interaction.
Piyathilaka~\etal.~\cite{Piyathilaka_2015_FSR} used 3D skeleton features and trained dynamic bayesian networks 
for domestic service robots. The output laser spot on object is also related to the work `a clickable world'~\cite{Nguyen_2008_RSJ}, which selects the appropriate behavior to execute for an assistive
object-fetching robot using the 3D location of the click by the laser pointer. However, it is challenging to directly use these approaches to detecting the forgotten actions and remind people. 




\section{Overview}\label{sec:ov}

We outline our approach in this section (see Fig.~\ref{fig:vrep}). The input to our system is \mbox{RGB-D} videos with the 3D joints of human skeletons from Kinect v2. We first decompose a video into a sequence of overlapping fixed-length temporal clips (step (1)). We then extract the human-skeleton-trajectory features and the interacting-object-trajectory features from the clips (introduced in Section.~\ref{sec:fea}). The human skeleton features and RGB-D object features have shown higher performance over RGB only features for the human action modeling~\cite{Koppula_2013_RSS,Wu_2014_CVPR,Lin_2014_CVPR}.

In order to build a compact representation of the action video, we draw parallels to document modeling in the natural language~\cite{Blei_2003_LDA} to represent a video as a sequence of words. We use $k$-means to cluster the human-skeleton-trajectories/interacting-object-trajectories from all the clips in the training set to form a \emph{human-dictionary} and an \emph{object-dictionary}, where we use the cluster centers as \emph{human-words} and \emph{object-words} ((2) in Fig.~\ref{fig:vrep}). Then, the video can be represented as a sequence of human-word and object-word indices by mapping its human-skeleton-trajectories/interacting-object-trajectories to the nearest human-words/object-words in the dictionary ((3) in Fig.~\ref{fig:vrep}). Also, an activity video is about a set of \emph{action-topics} indicating which actions are present in the video, and a set of \emph{object-topics} indicating which object types are interacting in the actions ((4) in Fig.~\ref{fig:vrep}).



We then build an unsupervised learning model that models the mapping of action-words/object-words to the action-topics/object-topics, as well as the co-occurrence and the temporal relations between the topics ((5) in Fig.~\ref{fig:vrep}).  
Using the learned model, we can assign the action-topic/object-topic to each clip. So the continuous clips with the same assigned action-topic form an action segment ((6) in Fig.~\ref{fig:vrep}).

The unsupervised action assignments of the clips are challenging because there is no annotation during the training stage. Besides extracting rich visual features, we further consider the relations among actions and objects. Unlike previous works, our model captures long-range relations between actions \eg, \emph{put-milk-back-to-fridge} is strongly related to \emph{fetch-milk-from-fridge} even with \emph{pour} and \emph{drink} between them. We model all pairwise co-occurrence and temporal casual relations between topics in a video, using a new probabilistic model (introduced in Section~\ref{sec:apc}). Specifically, we use a joint distribution as the correlated topic priors. They estimate which actions and objects are most likely to co-occur in a video. And we use a relative time distributions of topics to capture the temporal causal relations between actions, which estimate the possible temporal ordering of the occurring actions in the video. 



\section{Visual Features} \label{sec:fea}
We describe how we extract the visual features of a clip in this section. We extract both human-skeleton-trajectory features and the interacting-object-trajectory features from the output by the Kinect v2~\cite{kinect}
, which has an improved body tracker and the higher resolution of \mbox{RGB-D} frame than the Kinect v1. The tracked human skeleton has $25$ joints in total.  Let $X_u=\{x_u^{(1)},x_u^{(2)},\cdots,x_u^{(25)}\}$ be the 3D coordinates of $25$ joints of a skeleton in the current frame $u$.  We first compute the cosine of the angles between the connected body parts in each frame: $\alpha^{(pq)}=(p^{(p)}\cdot p^{(q)})/(|p^{(p)}|\cdot|p^{(q)}|)$, where the vector $p^{(p)}=x^{(i)}-x^{(j)}$ represents the body part. The transition between the joint coordinates and angles in different frames can well capture the human body movements. So we extract the motion features and off-set features~\cite{Wu_2014_CVPR} by computing their Euclidean distances $\D(,)$ to previous frame $f^m_{u,u-1}, f^\alpha_{u,u-1}$ and the first frame $f^m_{u,1},f^\alpha_{u,1}$ in the clip:
 \begin{equation}
 \begin{split}
 &f^m_{u,u-1}=\{\D(x^{(i)}_u,x^{(i)}_{u-1})\}_{i=1}^{25},\ f^\alpha_{u,u-1}=\{\D(\alpha^{(pq)}_u,\alpha^{(pq)}_{u-1})\}_{pq};\\
&f^m_{u,1}=\{\D(x^{(i)}_u,x^{(i)}_{1})\}_{i=1}^{25},\ f^\alpha_{u,1}=\{\D(\alpha^{(pq)}_u,\alpha^{(pq)}_{1})\}_{pq}.\notag
 \end{split}
 \end{equation}
 Then we concatenate all $f^m_{u,u-1},f^\alpha_{u,u-1},f^m_{u,1},f^\alpha_{u,1}$ as the human features of the clip.

 \begin{figure}[ht]
  \begin{center}
  \includegraphics[width=0.45\linewidth]{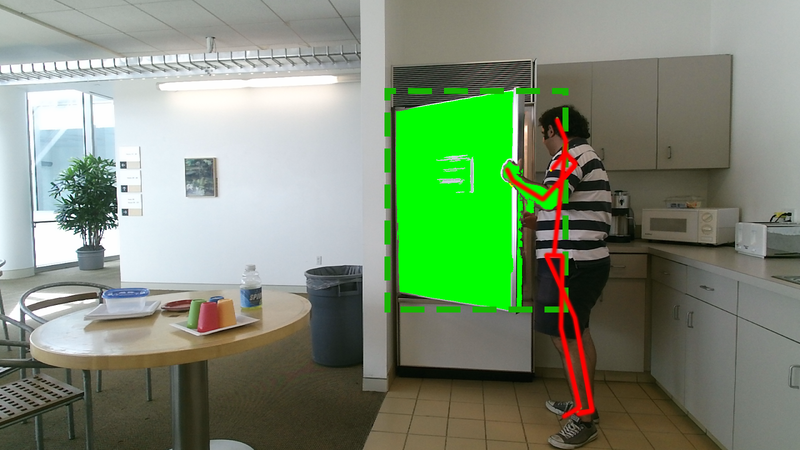}\hspace{0.05in}
  \includegraphics[width=0.45\linewidth]{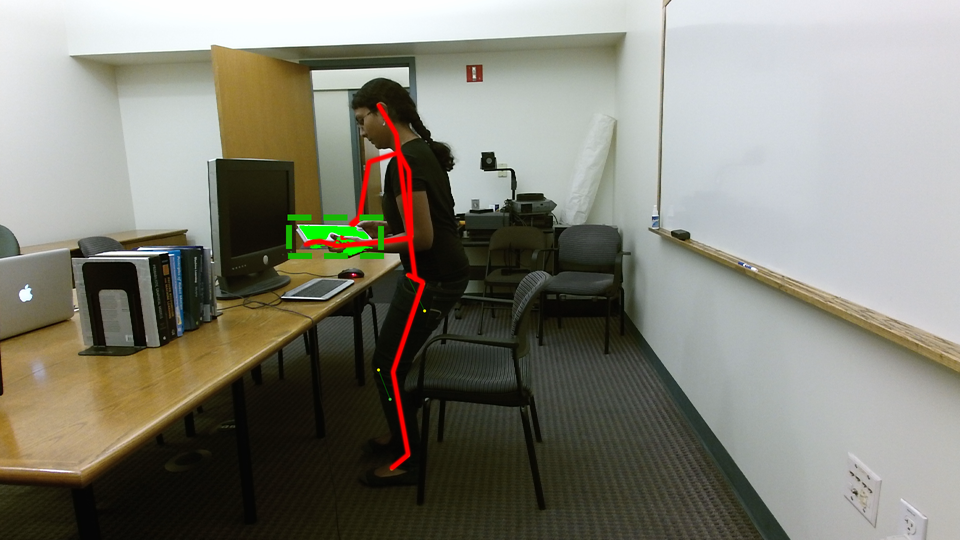}
 \caption{Examples of the human skeletons (red line) and the extracted interacting objects (green mask, left: fridge, right: book).} \label{fig:obj}
 \end{center}
\end{figure}

We also extract the human interacting-object-trajectory based on the human hands, image segmentation, motion detection and tracking. 
To detect the interacting objects, first we segment each frame into super-pixels using a fast edge detection approach~\cite{Dollar_2013_ICCV} on both RGB and depth images. 
The RGB-D edge detection provides richer candidate super-pixels rather than pixels to further extract objects. We then apply the moving foreground mask~\cite{Chris_CVPR_1999} to remove the unnecessary steady backgrounds and select those super-pixels within a distance to the human hands in both 3D points and 2D pixels. Finally, we collect the bounding boxes enclosing these super-pixels as the potential interested objects (see examples in Fig.~\ref{fig:obj}). 

We then track the bounding box in the segmented clip using SIFT matching and RANSAC to get the trajectories. We use the closest trajectory to the human hands for the clip. Finally, we extract six kernel descriptors~\cite{Ren_2012_CVPR} from the bounding box of each frame in the trajectory: gradient, color, local binary pattern, depth gradient, spin, surface normals, and KPCA/self-similarity, which have been proven to be useful features for \mbox{RGB-D} data~\cite{Wu_2014_RSS}. We concatenate the object features of each frame as the interacting-object-trajectory feature of the clip.

\section{Learning Model}\label{sec:apc}

\begin{figure}[t]
  \begin{center}
  \includegraphics[width=1\linewidth]{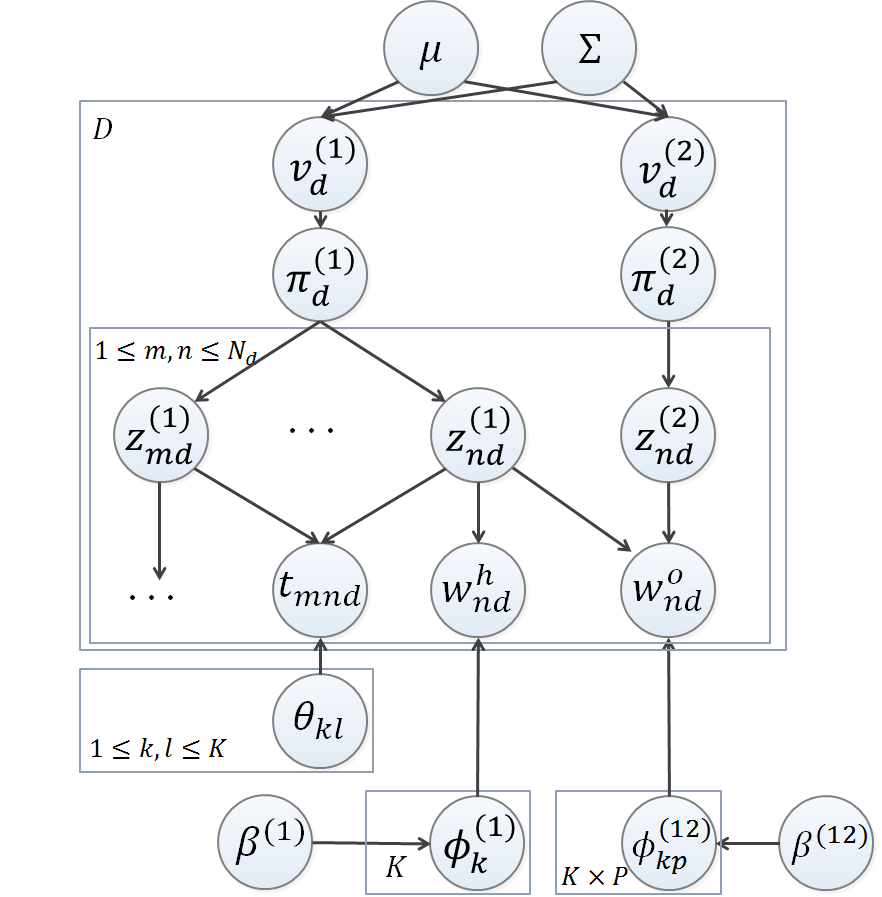}
 \caption{The graphic model of our causal topic model.} \label{fig:model}
 \end{center}
\end{figure}

In order to incorporate the aforementioned properties of activities, we
present a new generative model (see the graphic model in Fig.~\ref{fig:model} and the notations in Table~\ref{tb:not}). The novelty of our model is the ability to capture both short-range and long-range relations between actions in the compose activity videos in an unsupervised way. Using these relations, we can simultaneously segment the video and assign the action-topics as well as infer forgotten actions.

Consider a collection of $D$ videos (documents in the topic model). Each video as a document $d$ consists of $N_d$ continuous clips $\{c_{nd}\}_{n=1}^{N_d}$, each of which consists of a human-word $w^h_{nd}$ mapped to the human-dictionary and an object-word $w^o_{nd}$ mapped to the object-dictionary. 
We assign action-topic to each clip $c_{nd}$ from $K$ latent action-topics, indicating which action-topic they belong to. We assign object-topic to each object-word $w^o_{nd}$ from $P$ latent object-topics, indicating which object-topic is interacting within the clip. The assignments are denoted as $z^{(1)}_{nd}$ and $z^{(2)}_{nd}$. We use superscripts $(1),(2)$ to denote action-topics and object-topics respectively. After assignments, continuous clips with the same action-topic compose an action segment in a video. All the segments assigned with the same action-topic from the training set compose an action cluster.


The topic model such as LDA~\cite{Blei_2003_LDA} has been very common for document modeling from language. We use a it to generate a video document using a mixture of topics. Enable to model human actions in the video, our model introduces co-occurrence and temporal structure of topics instead of the topic independence assumption in LDA.

{\bf Basic generative process.} In a document $d$, we choose $z^{(1)}_{dn}\sim Mult(\pi^{(1)}_{:d}), z^{(2)}_{dn}\sim Mult(\pi^{(2)}_{:d})$, where $Mult(\pi)$ is a multinomial distribution with parameter $\pi$. The human-word $w^h_{nd}$ is drawn from an action-topic specific multinomial distribution $\phi^{(1)}_{z^{(1)}_{nd}}$, $w^h_{dn}\sim Mult(\phi^{(1)}_{z^{(1)}_{dn}})$, where $\phi^{(1)}_k\sim Dir(\beta^{(1)})$ is the human-word distribution of action-topic $k$, sampled from a Dirichlet prior with the hyperparameter $\beta^{(1)}$. While the object-word $w^o_{nd}$ is drawn from an action-topic and object-topic specific multinomial distribution $\phi^{(12)}_{z^{(1)}_{nd}z^{(2)}_{nd}}$, $w^o_{dn}\sim Mult(\phi^{(12)}_{z^{(1)}_{nd}z^{(2)}_{nd}})$, where $\phi^{(12)}_{kp}\sim Dir(\beta^{(12)})$ is the object-word distribution of action-topic $k$ and object-topic $p$. Here we consider the same object type like \emph{book} can be variant in appearance in different actions such as a close book in \emph{fetch-book} and a open book in \emph{read} action. So we consider the object-word distribution for different combinations of the action topic and the object topic.


{\bf Topic correlations.} The co-occurrence such as action \emph{pour} and action \emph{drink}, object \emph{book} and action \emph{read}, is useful to recognizing the co-occurring actions/objects and also gives a strong evidence for detecting forgotten actions. We model the co-occurrence by drawing their priors from a mixture distribution. Let $\pi^{(1)}_{kd}, \pi^{(2)}_{pd}$ be the probability of action-topic $k$ and object-topic $p$ occurring in document $d$, where $\sum_{k=1}^K\pi^{(1)}_{kd}=1, \sum_{p=1}^P\pi^{(2)}_{pd}=1$. Instead of sampling it from a fix Dirichlet prior with parameter in LDA that models them independently, we construct the probabilities by a stick-breaking process as follows. The stick-breaking notion has been widely used for constructing random weights~\cite{Sethuraman_1994_Statistica,Kim_2011_NIPS}.
\begin{equation}
\begin{split}
&\pi^{(1)}_{kd}=\Psi(v^{(1)}_{kd})\prod_{l=1}^{k-1}\Psi(v^{(1)}_{ld}),\ \Psi(v^{(1)}_{kd})=\frac{1}{1+\exp(-v^{(1)}_{kd})},\\
&\pi^{(2)}_{pd}=\Psi(v^{(2)}_{pd})\prod_{l=1}^{p-1}\Psi(v^{(2)}_{ld}),\ \Psi(v^{(2)}_{pd})=\frac{1}{1+\exp(-v^{(2)}_{pd})},\notag
\end{split}
\end{equation}
where $0<\Psi(v^{(1)}_{kd}), \Psi(v^{(2)}_{pd})<1$ is a classic logistic function, which satisfies $\Psi(-v^{(1)}_{kd})=1-\Psi(v^{(1)}_{kd}), \Psi(-v^{(2)}_{pd})=1-\Psi(v^{(2)}_{pd})$, and $v^{(1)}_{kd}, v^{(2)}_{pd}$ serves as the prior of $\pi^{(1)}_{kd}, \pi^{(2)}_{pd}$.

In order to capture the correlations between action-topics and object-topics, we draw the packed vector $v_{:d}=[v^{(1)}_{:d}, v^{(2)}_{:d}]$ in the stick-breaking notion from a mutivariate normal distribution $N(\mu,\Sigma)$. In practice, we use a truncated vector $v^{(1)}_{:d}=[v^{(1)}_{1d},\cdots,v^{(1)}_{K-1,d}]$ for (K-1) topics, and set $\pi^{(1)}_{Kd}=1-\sum_{k=1}^{K-1}\pi^{(1)}_{kd}=\prod_{k=1}^{K-1}\Psi(-v^{(1)}_{kd})$ as the probability of the final topic for a valid distribution. The same for $v^{(2)}_{:d}$.

 \begin{table}[t]
\footnotesize
\setlength{\tabcolsep}{1pt}
\caption{Notations in our model.}\label{tb:not}
\begin{tabular}{ll}
\hline
Symbols&Meaning\\
\hline
$D$ &number of videos in the training database;\\
$K$ &number of action-topics;\\
$P$ &number of object-topics;\\
$N_d$ &number of human-words/object-words in a video;\\
$c_{nd}$ &$n$-th clip in $d$-th video;\\
$w^h_{nd}$ &$n$-th human-word in $d$-th video;\\
$w^o_{nd}$ &$n$-th object-word in $d$-th video;\\
$z^{(1)}_{nd}$ &action-topic assignment of $c_{nd}$;\\
$z^{(2)}_{nd}$ &object-topic assignment of $w^o_{nd}$;\\
$t_{nd}$ &normalized timestamp of of $c_{nd}$;\\
$t_{mnd}$&$=t_{md}-t_{nd}$ the relative time between $c_{md}$ and $c_{nd}$;\\
$\pi^{(1)}_{:d}, \pi^{(2)}_{:d}$ &the probabilities of action/object-topics in $d$-th document;\\
$v^{(1)}_{:d}, v^{(2)}_{:d}$ &the priors of $\pi^{(1)}_{:d}, \pi^{(2)}_{:d}$ in $d$-th document;\\
$\phi^{(1)}_k$ &multinomial human-word distribution from action-topic $k$;\\
$\phi^{(12)}_{kp}$ &multinomial object-word distribution from \\
&action-topic $k$ and object-topic $p$;\\
$\mu,\Sigma$ &multivariate normal distribution of $v_{:d}=[v^{(1)}_{:d}, v^{(2)}_{:d}]$;\\
$\theta_{kl}$ &relative time distribution of $t_{mnd}$, between action-topic $k,l$;\\
\hline
\end{tabular}
\end{table}

\begin{figure}[t]
  \begin{center}
\includegraphics[width=0.9\linewidth]{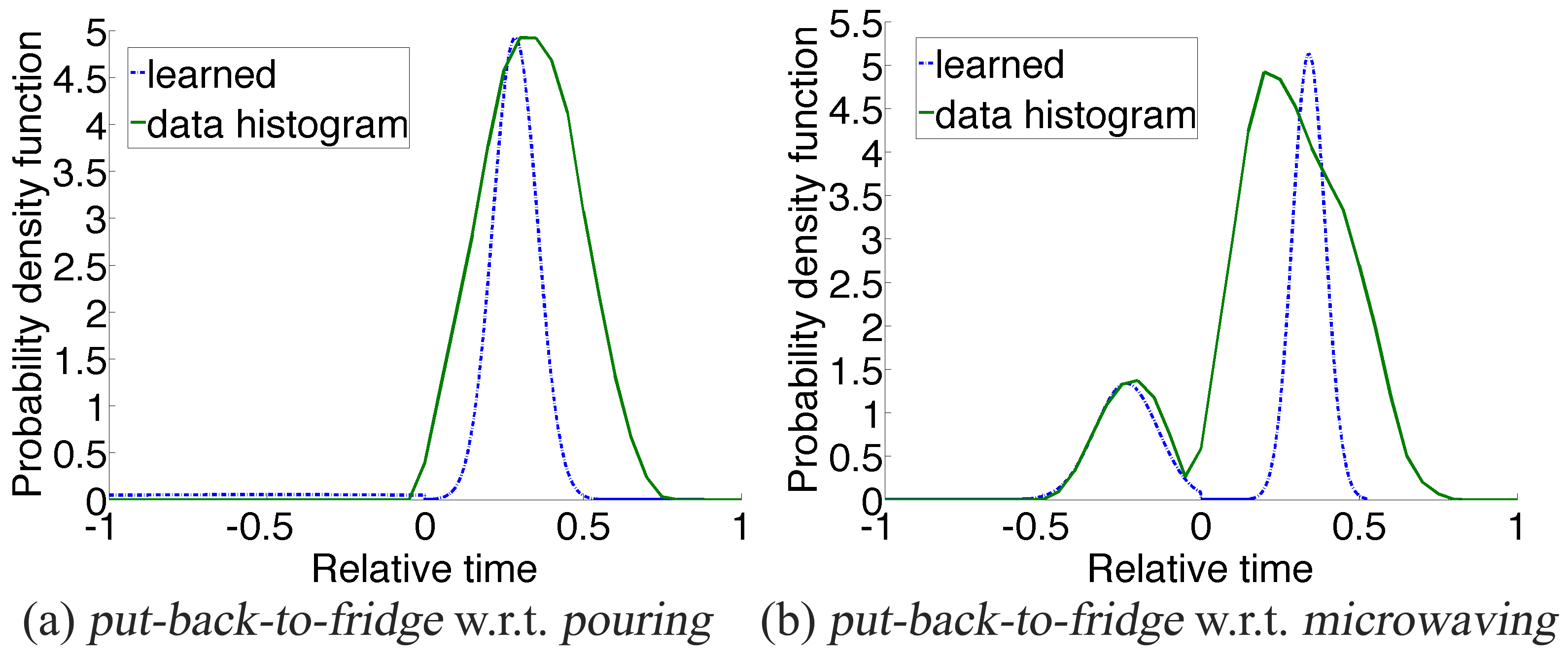}
 \caption{The relative time distributions learned by our model on training set (the blue dashed line) and the ground-truth histogram of the relative time over the whole dataset (the green solid line).} \label{fig:tpdf}
 \end{center}
\end{figure}
{\bf Relative time distributions.} The temporal relations between actions are also useful to discriminating the actions using temporal ordering and inferring the forgotten actions using the temporal context. We model the relative time of occurring actions by taking their time stamps into account. We consider that the relative time between two words are drawn from a certain distribution according to their topic assignments. In detail, let $t_{nd},t_{md}\in(0,1)$ be the absolute time stamp of $n$-th word and $m$-th word, which is normalized by the video length. $t_{mnd}=t_{md}-t_{nd}$ is the relative time of $m$-th clip relative to $n$-th clip. Then $t_{mnd}$ is drawn from a certain distribution, $t_{mnd}\sim \Omega(\theta_{z^{(1)}_{md},z^{(1)}_{nd}})$, where $\theta_{z^{(1)}_{md},z^{(1)}_{nd}}$ are the parameters. $\Omega(\theta_{k,l})$ are $K^{2}$ pairwise action-topic specific relative time distributions defined as follows:
\begin{equation}\label{eqn:tpdf}
\begin{split}
\Omega(t|\theta_{k,l})&=
\begin{cases}
b_{k,l}\cdot N(t|\theta^+_{k,l}) \ \ \ \ \ \ \ \ \ \ \textrm{if}\ \ t\geq 0,\\		
1-b_{k,l}\cdot N(t|\theta^-_{k,l}) \ \ \ \textrm{if}\ \ t<0,\\
\end{cases}
\end{split}
\end{equation}


An illustration of the learned relative time distributions are shown in Fig.~\ref{fig:tpdf}. We can see that the distributions we learned correctly reflect the order of the actions, \eg, \emph{put-back-to-fridge} is after \emph{pour} and can be before/after \emph{microwave}, and the shape is almost similar to the real distributions. Here the Bernoulli distribution $b_{k,l}/1-b_{k,l}$ gives the probability of action $k$ after/before the action $l$. And two independent normal distributions $N(t|\theta^+_{k,l})/N(t|\theta^-_{k,l})$ estimate how long the action $k$ is after/before the action $l$\footnote{Specially, when $k=l$, If two words are in the same segments, we draw $t$ from a normal distribution which is centered on zero, and the variance models the length of the action. If not, it also follows Eq.~(\ref{eqn:tpdf}) indicating the relative time between two same actions.  We also use functions $\tan(-\pi/2+\pi t) (0 < t < 1), \tan(\pi/2+\pi t) (-1 < t < 0)$ to feed $t$ to the normal distribution so that the probability is valid, that summits to one through the domain of $t$.}. Then the order and the length of the actions will be captured by all these pairwise relative time distributions.





\begin{figure*}[t]
  \begin{center}
  \subfigure[Robot System.]{
  \includegraphics[height=2.3in]{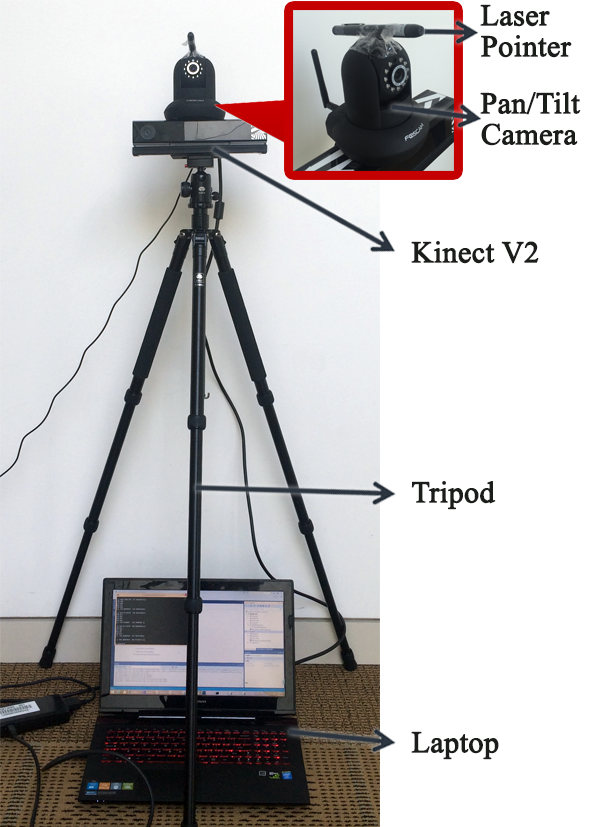}
  }
  \subfigure[System Pipeline.]{
  \includegraphics[height=2.3in]{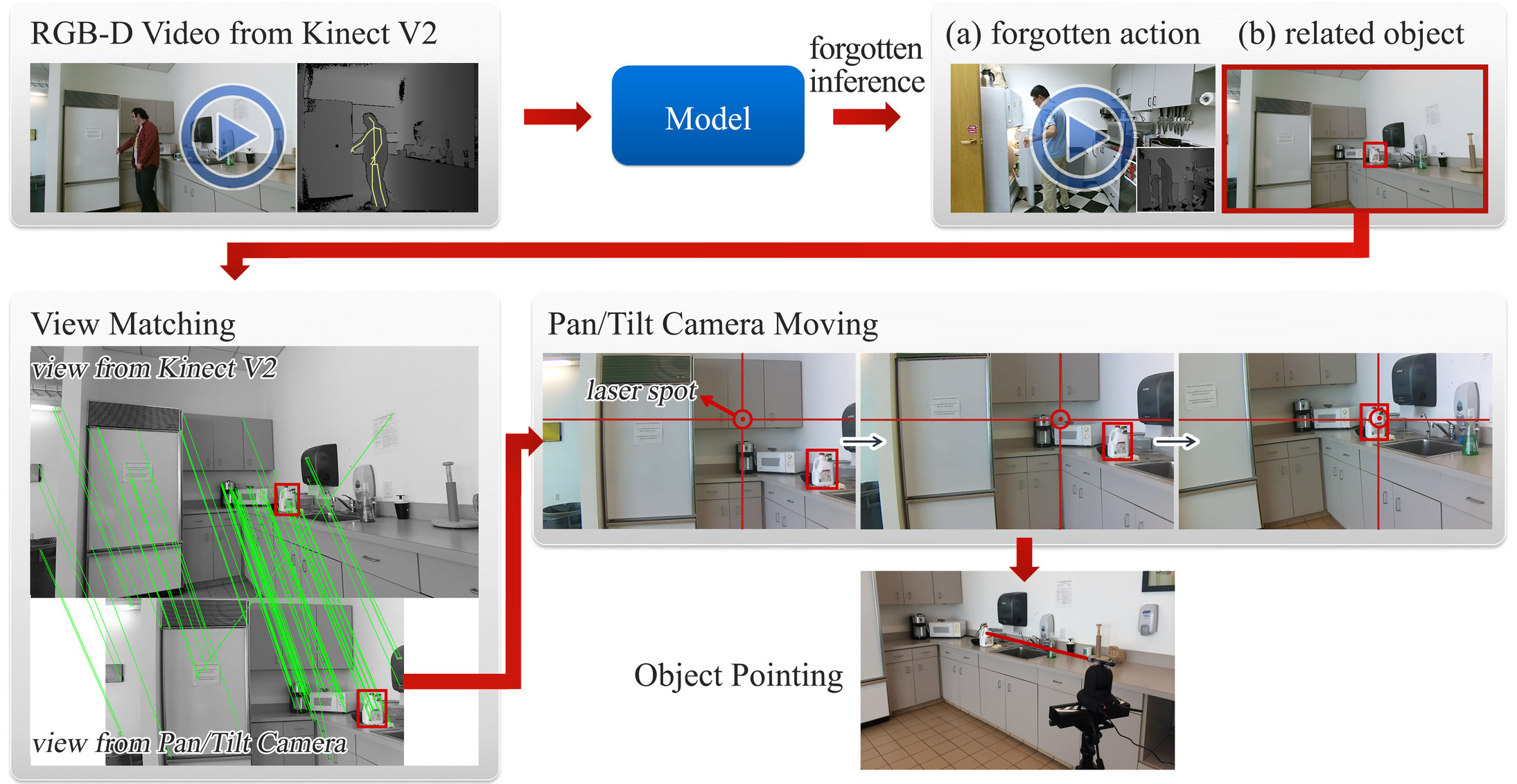}
  }
 \caption{ (a). Our Watch-Bot system. It consists of a Kinect v2 sensor that inputs \mbox{RGB-D} frames of human actions, a laptop that infers the forgotten action and the related object, a pan/tilt camera that localizes the object, mounted with a fixed laser pointer that points out the object. (b). The system pipeline.  The robot first uses the learned model to infer the forgotten action and the related object based on the Kinect's input. Then it maps the view from the Kinect to the pan/tilt camera so that the bounding box of the object is mapped in the camera's view. Finally, the camera moves until the laser spot lies in the bounding box of the target object.}
 \label{fig:system}
 \end{center}
\end{figure*}
\subsection{Learning and Inference}\label{sec:learn}
Gibbs sampling is commonly used as a means of statistical inference to approximate the distributions of variables when direct sampling is difficult~\cite{Blei_2009_TM,Kim_2011_NIPS}. Given a video, the word $w^h_{nd}, w^o_{nd}$ and the relative time $t_{mnd}$ are observed. We can integrate out $\Phi^{(1)}_k,\Phi^{(12)}_{kp}$ since $Dir(\beta^{(1)}), Dir(\beta^{(12)})$ are conjugate priors for the multinomial distributions $\Phi^{(1)}_k, \Phi^{(12)}_{kp}$. We also estimate the standard distributions including the mutivariate normal distribution $N(\mathbf{\mu},\Sigma)$ and the time distribution $\Omega(\theta_{kl})$ using the method of moments, once per iteration of Gibbs sampling. Following the convention, we use the fixed symmetric Dirichlet distributions by setting $\beta^{(1)}, \beta^{(12)}$ as $0.01$.

Then we introduce how we sample the topic assignment $z^{(1)}_{nd}, z^{(2)}_{nd}$. We do a collapsed sampling as in LDA by calculating the posterior distribution of $z^{(1)}_{nd}, z^{(2)}_{nd}$:
\vspace{-0.02in}
\begin{align}\label{eqn:posz}
&p(z^{(1)}_{nd}=k|\pi^{(1)}_{:d},z^{(1)}_{-nd},z^{(2)}_{nd},t_{nd})\notag\\
&\propto \pi^{(1)}_{kd}\omega(k,w^h_{nd})\omega(k,z^{(2)}_{nd},w^o_{nd})p(t_{nd}|z^{(1)}_{:d},\theta),\notag\\
&p(z^{(2)}_{nd}=p|\pi^{(2)}_{:d},z^{(2)}_{-nd},z^{(1)}_{nd})\propto
\pi^{(2)}_{pd}\omega(z^{(1)}_{nd},p,w^o_{nd}),\notag\\
&\omega(k,w^h_{nd})=\frac{N_{kw^h}^{-nd}+\beta^{(1)}}{N_{k}^{-nd}+N_{w^h}\beta^{(1)}},\notag\\
&\omega(k,p,w^o_{nd})=\frac{N_{kpw^o}^{-nd}+\beta^{(12)}}{N_{kp}^{-nd}+N_{w^o}\beta^{(12)}},\notag\\
&p(t_{nd}|z^{(1)}_{:d},\theta)=\prod_m^{N_d} \Omega(t_{mnd}|\theta_{z^{(1)}_{md},k})\Omega(t_{nmd}|\theta_{k,z^{(1)}_{md}}),
\vspace{-0.05in}
\end{align}
where $N_{w^h},N_{w^o}$ is the number of unique word types in dictionary, $N_{kw^h}^{-nd}/N_{kpw^o}^{-nd}$ denotes the number of instances of word $w^h_{nd}/w^o_{nd}$ assigned with action-topic $k$/action-topic $k$ and object-topic $p$, excluding $n$-th word in $d$-th document, and $N_{k}^{-nd}/N_{kp}^{-nd}$ denotes the number of total words assigned with action-topic $k$/action-topic $k$ and object-topic $p$. $z^{(1)}_{-nd}/z^{(2)}_{-nd}$ denotes the topic assignments for all words except $z^{(1)}_{nd}/z^{(2)}_{nd}$. The detailed derivation of Eq.~(\ref{eqn:posz}) is in the Appendix \ref{app:dev}.

In Eq.~(\ref{eqn:posz}), note that the topic assignments are decided by which actions/objects are more likely to co-occur in the video (the occurance probabilities $\pi^{(1)}_{kd}/\pi^{(2)}_{kd}$), the visual appearance of the word (the word distributions $\omega(k,w^h_{nd}), \omega(k,p,w^o_{nd})$) and the temporal relations (the relative time distributions $p(t_{nd}|z^{(1)}_{:d},\theta)$). 



Due to the logistic stick-breaking transformation, the posterior distribution of the topic priors $v_{:d}=[v^{(1)}_{:d}, v^{(2)}_{:d}]$ does not have a closed form. So we instead use a Metropolis-Hastings independence sampler~\cite{Gelman_2013_bayesian}. Let the proposals $q(v_{:d}^*|v_{:d},\mu,\Sigma)=N(v_{:d}^*|\mu,\Sigma)$ be drawn from the prior. The proposal is accepted with probability $\min(\A(v_{:d}^*,v_{:d}),1)$, where
 \begin{equation}
 \begin{split}
&\A(v_{:d}^*,v_{:d})\\
&=\frac{p(v_{:d}^*|\mu,\Sigma)\prod_{n=1}^{N_d}p(z^{(1)}_{nd}|v_{:d}^{(1)*})p(z^{(2)}_{nd}|v_{:d}^{(2)*})
q(v_{:d}|v_{:d}^*,\mu,\Sigma)}{p(v_{:d}|\mu,\Sigma)\prod_{n=1}^{N_d}p(z^{(1)}_{nd}|v^{(1)}_{:d})p(z^{(2)}_{nd}|v^{(2)}_{:d})q(v_{:d}^*|v_{:d},\mu,\Sigma)}\\
&=\prod_{n=1}^{N_d}\frac{p(z^{(1)}_{nd}|v_{:d}^{(1)*})p(z^{(2)}_{nd}|v_{:d}^{(2)*})}{p(z^{(1)}_{nd}|v^{(1)}_{:d})p(z^{(2)}_{nd}|v^{(2)}_{:d})}\\
&=\prod_{k=1}^K(\frac{\pi^{(1)*}_{kd}}{\pi^{(1)}_{kd}})^{\sum_{n=1}^{N_d}\delta(z^{(1)}_{nd},k)}
\prod_{p=1}^P(\frac{\pi^{(2)*}_{pd}}{\pi^{(2)}_{pd}})^{\sum_{n=1}^{N_d}\delta(z^{(2)}_{nd},p)},\notag
 \end{split}
 \end{equation}
 which can be easily calculated by counting the number of words assigned with each topic by $z^{(1)}_{nd}, z^{(2)}_{nd}$. Here the function $\delta(x,y)=1$ if only if $x=y$, otherwise equal to $0$. The time complexity of the sampling per iteration is $O(N_dD(\max(N_dK,P)))$

For inference of a test video, we sample the unknown topic assignments $z^{(1)}_{nd}, z^{(2)}_{nd}$ and the topic priors $v^{(1)}_{:d},v^{(2)}_{:d}$ using the learned parameters in the training stage.



 \section{Watch-Bot to Reminding of Forgotten Actions}\label{sec:ap}
 The average adult forgets three key facts, chores or events every day~\cite{forget}. So it is important for a personal robot to be able to detect not only what a human is currently doing but also what he forgot to do. In this section, we describe a new robot system (see Fig.~\ref{fig:system}) to detect the forgotten actions and remind people, which we called \emph{action patching}, using our learning model. 

Note that detecting forgotten action is more challenging than conventional action recognition, since what to infer is not shown in the query video. Also, our model does not necessarily know the semantic class of the actions. Instead it learns action clusters and relations from the unlabeled action videos and use them to detect forgotten actions and remind people. Therefore, modeling rich relations from videos is important to providing evidence for detecting forgotten actions. Our model models pairwise co-occurrence and long-range temporal relations of actions/topics. As a result, rather than only modeling the single action or the local temporal transitions in the previous works, those actions occurred with a relatively large time interval, occurred after the forgotten actions, as well as the interacting objects can also be used to detect forgotten actions in our model. For example, a \emph{put-back-book} might be forgotten as previously seen a \emph{fetch-book} action before a long \emph{read} action, and seen a \emph{book} and a \emph{leave} action indicates he really forgot it.
 
We enable a robot, that we call Watch-Bot, to detect humans' forgotten actions as well as localize the related object in the current scene. The robot consists of a Kinect v2 sensor, a pan/tilt camera (which we call camera for brevity in this paper) mounted with a laser pointer, and a laptop (see Fig.~\ref{fig:system}). This setup can be easily deployed on any assistive robot. Taking the example in Fig.~\ref{fig:tr}, if our robot sees a person fetch a milk from the fridge, pour the milk, and leave without putting the milk back to the fridge. Our robot would first detect the forgotten action and the related object (the milk), given the input \mbox{RGB-D} frames and human skeletons from the Kinect; then map the object from the Kinect's view to the camera's view; finally pan/tilt the camera till its mounted laser pointer pointing to the milk.


Our goal is to detect the forgotten action and then point out the related object in the forgotten action using our learned model (see Alg.~\ref{alg:adr}). We first use our model to segment the query video into action segments (step 1,2 in Alg.~\ref{alg:adr}), and then infer the most possible forgotten action-topic and the related object-topic (step 4 in Alg.~\ref{alg:adr}). Next we retrieve a top forgotten action segment from the training database, containing the inferred forgotten action-topic and the object-topic (step 5,6 in Alg.~\ref{alg:adr}). Using the extracted object in the retrieved segment, we detect the bounding box of the related forgotten object in the Kinect's view of the query video (step 8,9,10 in Alg.~\ref{alg:adr}). After that, we map the bounding box of the object from the Kinect's view to the camera's view. Finally, the pan/tilt camera moves until its mounted laser pointer points out the related object in the current scene.


{\bf Patched Action and Object Inference.} Our model infers the forgotten action using the probability inference based on the dependencies between actions and objects. After assigning the action-topics and object-topics to a query video $q$, we consider adding one additional clip $\hat{c}$ consisting of $\hat{w^h}, \hat{w^o}$ into $q$ in each action segmentation point $t_s$ (see Fig~\ref{fig:vp}). Then the probabilities of the missing action-topics $k_m$ with object-topics $p_m$ in each segmentation point $t_s$ can be compared following the posterior distribution in Eq.~(\ref{eqn:posz}):
\begin{align}\label{eqn:patch}
&p(z^{(1)}_{\hat{c}}=k_m, z^{(2)}_{\hat{c}}=p_m, t_{\hat{c}}=t_s|other)\notag\\
&\propto \pi^{(1)}_{k_md}\pi^{(2)}_{p_md}p(t_{s}|z^{(1)}_{:d},\theta)\sum_{w^h,w^o}\omega(k_m,w^{h})\omega(k_m,p_m,w^{o}),\notag\\
&\st \ \ \ t_s\in{T_s},\ k_m\in{[1:K]}-K_e,
\end{align}
where $T_s$ is the set of segmentation points ($t_1,t_2$ in Fig.~\ref{fig:vp}) and $K_e$ is the set of existing action-topics in the video (\emph{fetch-book}, \etc in Fig.~\ref{fig:vp}). Thus ${[1:K]}-K_e$ are the missing topics in the video (\emph{put-down-items}, \etc in Fig.~\ref{fig:vp}). $p(t_{s}|z^{(1)}_{:d},\theta),\omega(k_m,w^{h}),\omega(k_m,p_m,w^{o})$ can be computed as in Eq.~(\ref{eqn:posz}). Here we marginized $\hat{w^h},\hat{w^o}$ to avoid the effect of a specific human-word or object-word. Note that, $\pi^{(1)}_{kd}, \pi^{(2)}_{pd}$ gives the probability of a missing action-topic with an object-topic in the video decided by the correlation we learned in the joint distribution prior, \ie, the close topics have higher probabilities to occur in this query video. And $p(t_{s}|z^{(1)}_{:d},\theta)$ measures the temporal consistency of adding a new action-topic. And the marginized word-topic distribution $\sum_{w^h,w^o}\omega(k_m,w^{h})\omega(k_m,p_m,w^{o})$ give the likelihood of the topic learned from training data.

{\setlength{\textfloatsep}{2pt}
\begin{algorithm}[t]
\small
\caption{Forgotten Action and Object Detection.}
\begin{algorithmic}
\STATE \textbf{Input:} RGB-D video $q$ with tracked human skeletons.
\STATE \textbf{Output:} Claim no action forgotten, or output an action segment with the forgotten action and a bounding box of the related object in the current scene.
\STATE 1. Assign the action-topics to clips and the object-topics to object-words in $q$ as introduced in Section~\ref{sec:learn}.
\STATE 2. Get the action segments by merging the continuous clips with the same assigned action-topic.
\STATE 3. If the assigned action-topics $K_e$ in $q$ contains all modeled action-topics $[1:K]$, claim no action forgotten and return;
\STATE 4. For each action segmentation point $t_s$, each not assigned action-topic $k_m\in{[1:K]}-K_e$, and each object-topic $p_m\in{[1:P]}$:
\STATE \ \ \ Compute the probability defined in Eq.~\ref{eqn:patch};
\STATE 5. Select the top tree possible tuples $(k_m,p_m,t_s)$, and get the forgotten action segment candidate set $Q$ which contains segments with topics $(k_m,p_m)$;
\STATE 6. Select the top forgotten action segment $p$ from $Q$ with the maximum $patch\_score(p)$;
\STATE 7. If $patch\_score(p)$ is smaller than a threshold, claim no action forgotten and return;
\STATE 8. Segment the current frame to super-pixels using edge detection~\cite{Dollar_2013_ICCV} as in Section~\ref{sec:ov};
\STATE 9. Select the nearest super-pixels to both extracted object bounding box in $q$ and $p$.
\STATE 10. Merge the adjacent super-pixels and bound the largest one with a rectangle as the output bounding box.
\STATE 11. Return the top forgotten action segment and the object bounding box.
\end{algorithmic}  \label{alg:adr}
\end{algorithm}

{\bf Patched Action and Object Detection.} Then we select the top three tuples $(k_m,p_m,t_s)$ using the above probability. The action segments of action-topic $k_m$ containing object-topic $p_m$ in the training set consist a patched action candidate segment set $Q$. We then select the patched action segment from $Q$ with the maximum $patch\_score$ defined in Eq.~\ref{eqn:rank}. In detail, we consider that the front and the tail of the patched action segment $f_{pf}, f_{pt}$ should be similar to the tail of the adjacent segment in $q$ before $t_s$ and the front of the adjacent segment in $q$ after $t_s$: $f_{qt}, f_{qf}$. At the same time, the middle of the patched action segment $f_{pm}$ should be different to $f_{qt},f_{qf}$, as it is a different action forgotten in the video.\footnote{Here the middle, front, tail frames are $20\%$-length of segment centering on the middle frame, starting from the first frame, and ending at the last frame in the segment respectively.}
\begin{equation}\label{eqn:rank}
\begin{split}
patch\_score(p) = ave(\D(f_{pm},f_{qf}),\D(f_{pm},f_{qt}))\\
-max(\D(f_{pf},f_{qt}),\D(f_{pt},f_{qf})),
\end{split}
\end{equation}
where $\D(,)$ is the average pairwise distances between frames, $ave(,),max(,)$ are the average and max value. If the maximum score is below a threshold or there is no missing topics $(\ie, K_e=[1:K])$ in the query video, we claim there is no forgotten actions. Then we detect the bounding box of the patched object. We first segment the current frame into super-pixels as in Section~\ref{sec:ov}, second search the nearest segments using the extracted object in the test video and the patched action, finally merge the adjacent segments into one segment and bound the largest segment with a bounding box.

\begin{figure}[t]
  \begin{center}
  \includegraphics[width=0.9\linewidth]{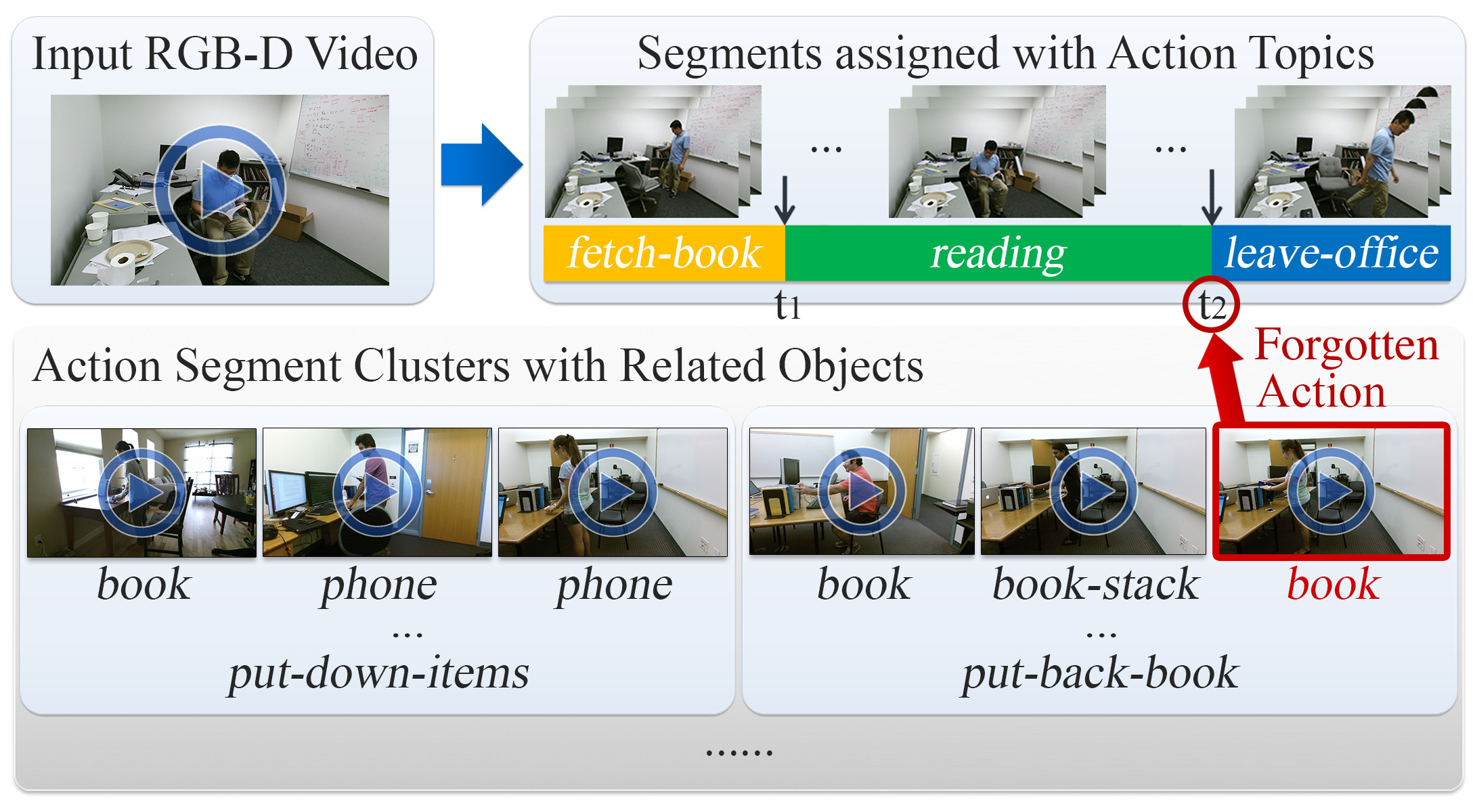}
  \caption{{\bf Illustration of patched action and object inference using our model.} Given a test video, we infer the forgotten action-topic and object-topic in each segmentation point ($t_1,t_2$ as above). Then we select the top segment from the inferred action-topic's segment cluster with the inferred object-topic with the maximum $patch\_score$.}
   \label{fig:vp}
 \end{center}
\end{figure}

{\bf Real Object Pointing.} We now describe  how we pan/tilt the camera to point out the real object in the current scene. We first compute the transformation homography matrix between the frame of the Kinect and the frame of the pan/tilt camera using keypoints matching and RANSAC, which can be done very fast within $0.1$ second. Then we can transform the detected bounding box from the Kinect's view to the pan/tilt camera's view. Since the position of the laser spot in the pan/tilt camera view is fixed, next we only need to pan/tilt the camera till the laser spot lies within the bounding box of the target object. To avoid the coordinating error caused by distortion and inconsistency of the camera movement, we use an iterative search plus small step movement instead of one step movement to localize the object (illustrated in Fig.~\ref{fig:system}). In each iteration, the camera pan/tilt a small step towards to the target object according to the relative position between the laser spot and the bounding box. Then the homography matrix is recomputed in the new camera view, so that the bounding box is mapped in the new view. Until the laser spot is close enough to the center of the bounding box, the camera stops moving.


\section{Experiments}\label{sec:exp}
\subsection{Watch-n-Patch Dataset}
We collect a new challenging \mbox{RGB-D} activity dataset recorded by the new Kinect v2 camera. Each video in the dataset contains $2$-$7$ actions interacting with different objects (see examples in Fig.~\ref{fig:actions}). The new Kinect v2 has higher resolution of \mbox{RGB-D} frames (RGB: $1920\times 1080$, depth: $512\times 424$) and improved body tracking of human skeletons ($25$ body joints). We record $458$ videos with a total length of about $230$ minutes. We ask $7$ subjects to perform human daily activities in $8$ offices and $5$ kitchens with complex backgrounds. And in each environment the activities are recorded in different views. It composed of fully annotated $21$ types of actions ($10$ in the office, $11$ in the kitchen) interacting with $23$ types of objects. We also record the audio, though it is not used in this paper.

In order to get a variation in activities, we ask participants to finish task with different combinations of actions and ordering. Some actions occur together often such as \emph{fetch-from-fridge} and \emph{put-back-to-fridge} while some are not always in the same video such as \emph{take-item} and \emph{read}. Some actions are in fix ordering such as \emph{fetch-book} and \emph{put-back-book} while some occur in random order such as \emph{put-back-to-fridge} and \emph{microwave}. Moreover, to evaluate the action patching performance, $222$ videos in the dataset has action forgotten by people and the forgotten actions are annotated. We give the examples of action classes in Fig.~\ref{fig:actions} and action sequences in Table~\ref{tb:data}.


\subsection{Experimental Setting and Compared Baselines}
We evaluate in two environments `office' and `kitchen'. In each environment, we split the data into a train set with most full videos (office: $87$, kitchen $119$) and a few forgotten videos (office: $10$, kitchen $10$), and a test set with a few full videos (office: $10$, kitchen $20$) and most forgotten videos (office: $89$, kitchen $113$). In our experiments, we compare seven unsupervised approaches with only action-topics. They are Hidden Markov Model (HMM), topic model LDA (TM), correlated topic model (CTM), topic model over absolute time (TM-AT), correlated topic model over absolute time (CTM-AT), topic model over relative time (TM-RT) and our causal topic model with only action-topics (CaTM-A)~\cite{Wu_2015_CVPR}. We compare three methods with both action-topics and object-topics. They are HMM with the object-topics (HMM-O), LDA with the object-topics (TM-O) and our causal topic model with the object-topics (CaTM-AO). All these methods use the same human skeleton and \mbox{RGB-D} features introduced in Section~\ref{sec:fea}. We also evaluate HMM and our model CaTM using the popular features for action recognition, dense trajectories feature (DTF)~\cite{Wang_2011_CVPR}, extracted only in RGB videos\footnote{We train a codebook with the size of $2000$ and encode the extracted DTF features in each clip as the bag of features using the codebook.}, named as HMM-DTF and CaTM-A-DTF, CaTM-AO-DTF.

In the experiments, we set the number of topics and states of HMM equal to or more than ground-truth classes. For correlated topic models, we use the same topic prior in our model. For models over absolute time, we consider the absolute time of each word is drawn from a topic-specific normal distribution.  For models over relative time, we use the same relative time distribution as in our model (Eq.~(\ref{eqn:tpdf})). The clip length of the action-words is set to $20$ frames, densely sampled by step one and the size of action dictionary is set to $500$. For patching, the candidate set for different approaches consist of the segments with the inferred missing topics by transition probabilities for HMM, the topic priors for TM and CTM, and both the topic priors and the time distributions for TM-AT, TM-RT, CTM-AT and our CaTM. Then we use the same $patch\_score$ as in Section~\ref{sec:ap} to select the top one patched segments, and the average of the $patch\_score$ computed in a set of the segmented videos after training is set as the threshold of claiming forgotten action.

\begin{table}[t]
\footnotesize
\setlength{\tabcolsep}{1pt}
\begin{center}
\caption{Results using the same number of topics as the ground-truth action classes. HMM-DTF, CaTM-A-DTF, CaTM-AO-DTF use DTF RGB features and others use our human skeleton and \mbox{RGB-D} features. (top one is bold)}\label{tb:re}
\begin{tabular*}{\linewidth}{@{\extracolsep{\fill}}c|c c |c c |c c| c c}
\hline
`office'&\multicolumn{2}{c|}{Seg-Acc}&\multicolumn{2}{c|}{Seg-AP}&\multicolumn{2}{c|}{Frame-Acc}&PA-Acc&PO-Acc\\
(\%)&Offline&Online&Offline&Online&Offline&Online&\\
\hline\hline
HMM-DTF&15.2&9.4&21.4&20.7&20.2&15.9&23.6&-\\
HMM&18.0&14.0&25.9&24.8&24.7&21.3&33.3&-\\
HMM-O&18.2&19.4&26.2&23.1&25.3&27.3&32.2&20.4\\
TM&9.3&9.2&20.9&19.6&20.3&13.0&13.3&-\\
TM-O&9.8&12.2&22.3&19.6&24.6&18.4&15.7&10.5\\
CTM&10.0&5.9&18.1&15.8&20.2&16.4&13.3&-\\
TM-AT&8.9&3.7&25.4&19.0&18.6&13.8&12.0&-\\
CTM-AT&9.6&6.8&25.3&19.8&19.6&15.5&10.8&-\\
TM-RT&30.8&30.9&29.0&30.2&38.1&36.4&39.5&-\\
CaTM-A-DTF&28.2&27.0&28.3&27.4&37.4&34.0&33.7&-\\
CaTM-AO-DTF&28.5&29.1&30.6&29.5&37.9&35.0&36.2&30.5\\
CaTM-A&30.6&32.9&\textbf{33.1}&34.6&39.9&38.5&41.5&-\\
CaTM-AO&\textbf{33.2}&\textbf{35.2}&33.0&\textbf{36.0}&\textbf{40.1}&\textbf{41.2}&\textbf{46.2}&\textbf{36.4}\\
\hline
\hline
`kitchen' &\multicolumn{2}{c|}{Seg-Acc}&\multicolumn{2}{c|}{Seg-AP}&\multicolumn{2}{c|}{Frame-Acc}&PA-Acc&PO-Acc\\
(\%)&Offline&Online&Offline&Online&Offline&Online&\\
\hline\hline
HMM-DTF&4.9&3.6&18.8&5.6&12.3&9.8&2.3&-\\
HMM&20.3&15.2&20.7&13.8&21.0&18.3&7.4&-\\
HMM-O&23.9&17.2&21.1&18.8&23.5&20.3&12.4&5.3\\
TM&7.9&4.7&21.5&14.7&20.9&11.5&9.6&-\\
TM-O&7.9&6.7&22.6&17.1&24.9&14.4&10.8&5.3\\
CTM&10.5&9.2&20.5&14.9&18.9&15.7&6.4&-\\
TM-AT&8.0&4.8&21.5&21.6&20.9&14.0&7.4&-\\
CTM-AT&9.7&10.0&19.1&22.6&20.1&16.7&10.7&-\\
TM-RT&32.3&26.9&23.4&23.0&35.0&31.2&18.3&-\\
CaTM-A-DTF&26.9&23.6&18.4&17.4&33.3&29.9&16.5&-\\
CaTM-AO-DTF&27.2&25.3&19.1&18.6&32.9&30.2&17.6&13.2\\
CaTM-A&\textbf{33.2}&29.0&26.4&25.5&37.5&34.0&20.5&-\\
CaTM-AO&32.1&\textbf{30.7}&\textbf{28.5}&\textbf{28.5}&\textbf{39.2}&\textbf{36.9}&\textbf{24.4}&\textbf{20.6}\\
\hline
\end{tabular*}
\end{center}
\end{table}

\subsection{Evaluation Metrics}

{\bf Action Segmentation and Cluster Assignment.} We want to evaluate if the unsupervised learned action-topics and states of HMM are semantically meaningful. In the unsupervised setting, we need to map the assigned topics to the ground-truth labels for evaluation. This could be done by counting the mapped frames between topics and ground-truth classes. Let $k_i,c_i$ be the assigned topic and ground-truth class of frame $i$. The count of a mapping is:
$m_{kc}=\frac{\sum_i \delta(k_i,k)\delta(c_i,c)}{\sum_i\delta(c_i,c)}$, where $\sum_i \delta(k_i,k)\delta(c_i,c)$ is the number of frames assigned with topic $k$ as the ground-truth class $c$ and normalized by the number of frames as the ground-truth class $c$: $\sum_i\delta(c_i,c)$. Then we can solve the following binary linear programming to get the best mapping:
\begin{equation}
\begin{split}
&\max_{x}\sum_{k,c}x_{kc}m_{kc},\\
s.t.\ \ \  \forall k,\ \sum_cx_{kc}=1,&\ \ \ \ \forall c,\ \sum_k x_{kc}\geq 1,\ \ \ \ x_{kc}\in \{0,1\},\notag
\end{split}
\end{equation}
where $x_{kc}=1$ indicates mapping topic $k$ to class $c$, otherwise $x_{kc}=0$. And $\sum_cx_{kc}=1$ constrain that each topic must be mapped to exact one class, $\sum_k x_{kc}\geq 1$ constrain that each class must be mapped by at least one topic.

We then measure the performance in two ways. Per frame: we compute \emph{frame-wise accuracy (Frame-Acc)}, the ratio of correctly labeled frames. Segmentation: we consider a true positive if the overlap (union/intersection) between the detected and the ground-truth segments is more than a default threshold $40\%$ as in~\cite{Pirsiavash_2014_CVPR}. Then we compute \emph{segmentation accuracy (Seg-Acc)}, the ratio of the ground-truth segments that are correctly detected, and \emph{segmentation average precision (Seg-AP)} by sorting all action segments output by the approach using the average probability of their words' topic assignments. All above three metrics are computed by taking the average of each action class.

{\bf Forgotten Action and Object Detection.} We also evaluate the \emph{patching accuracy (PA-Acc)} by the portion of correct patched video, including correctly output the forgotten action segments or correctly claiming no forgotten actions. We consider the output action segments by the algorithm containing over $50\%$ ground-truth forgotten actions as correctly output the forgotten action segments. We also measure the \emph{patching object detection accuracy (PO-Acc)} by the typical object detection metric,  that considers a true positive if the overlap rate (union/intersection) between the detected and the ground-truth object bounding box is greater than $40\%$.




\subsection{Results}

Table~\ref{tb:re} and Fig.~\ref{fig:act} show the main results of our experiments. We first
perform evaluation in the offline setting
to see if actions can be well segmented and clustered in the train set.
We then perform testing in an online setting to see if the new video from the test set can be correctly segmented and the segments can be correctly assigned to the action cluster. We can see that our approach performs better than the state-of-the-art in unsupervised action segmentation and clustering, as well as action patching. We discuss our results in the light of the following questions.

\begin{figure}[t]
  \begin{center}\hspace{-0.1in}
  \includegraphics[width=1\linewidth]{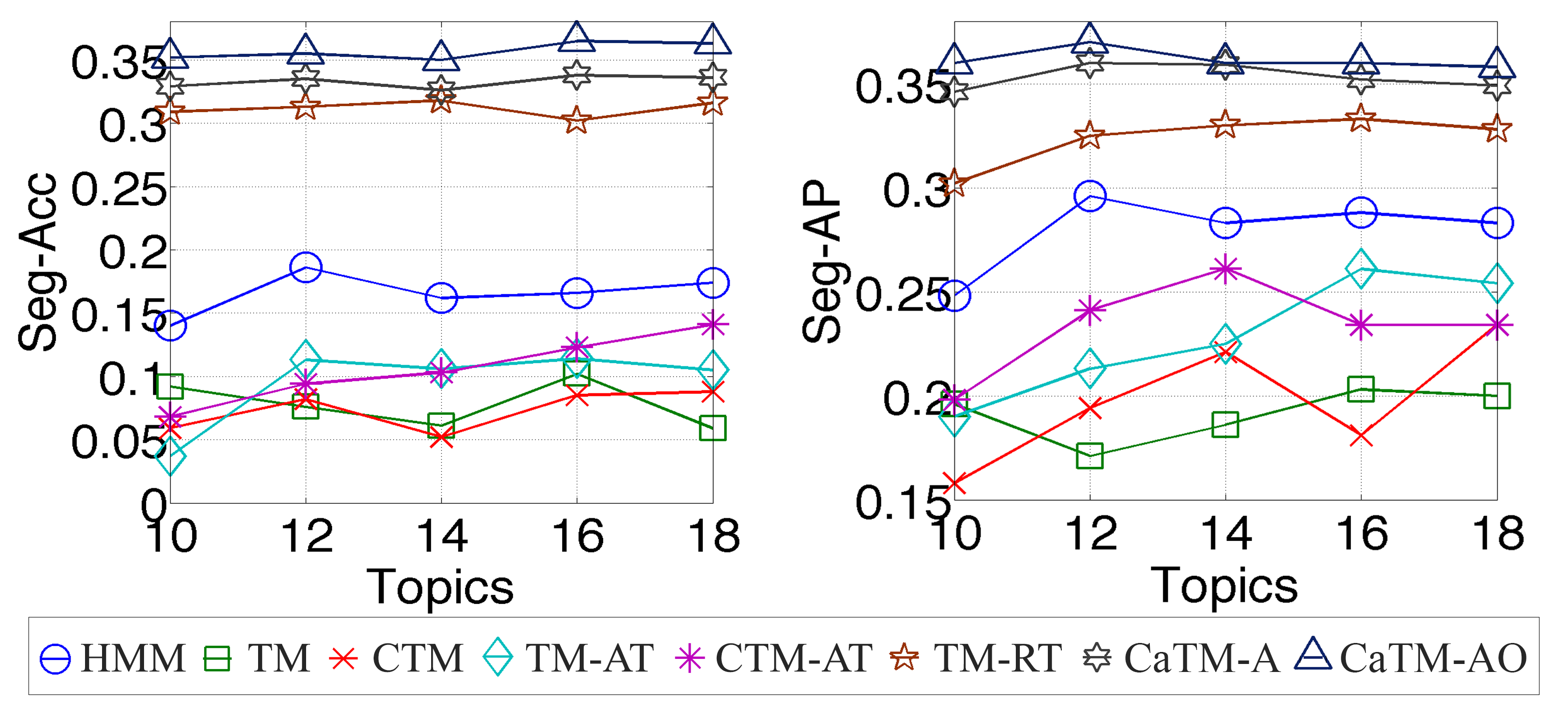}
 \hspace{-0.1in}
 \caption{Online segmentation Acc/AP varied with the number of topics in the `office' dataset.} \label{fig:act}
 \end{center}
\end{figure}

\begin{figure}[t]
  \begin{center}\hspace{-0.3in}
  \includegraphics[width=0.54\linewidth]{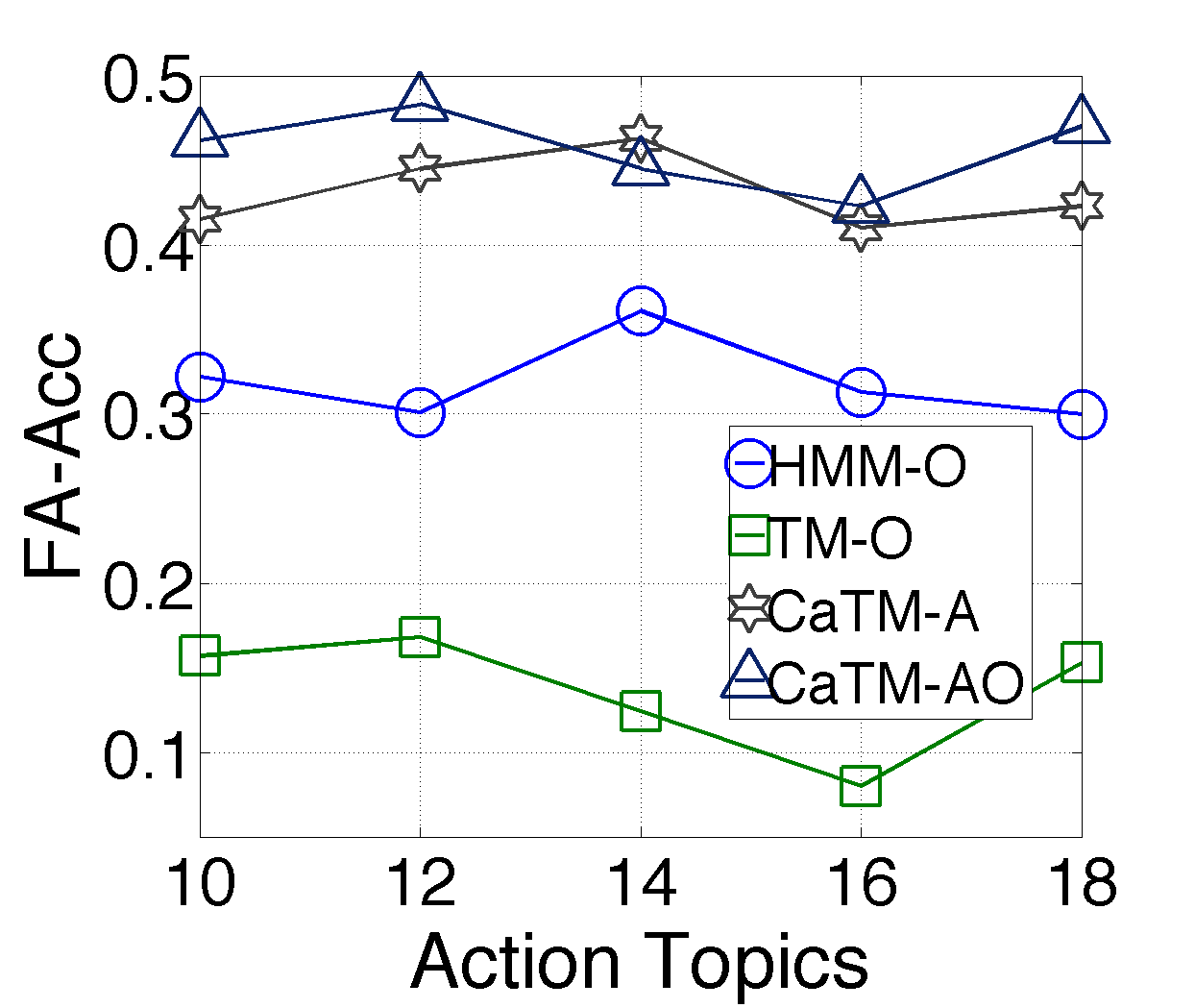}\hspace{-0.15in}
  \includegraphics[width=0.54\linewidth]{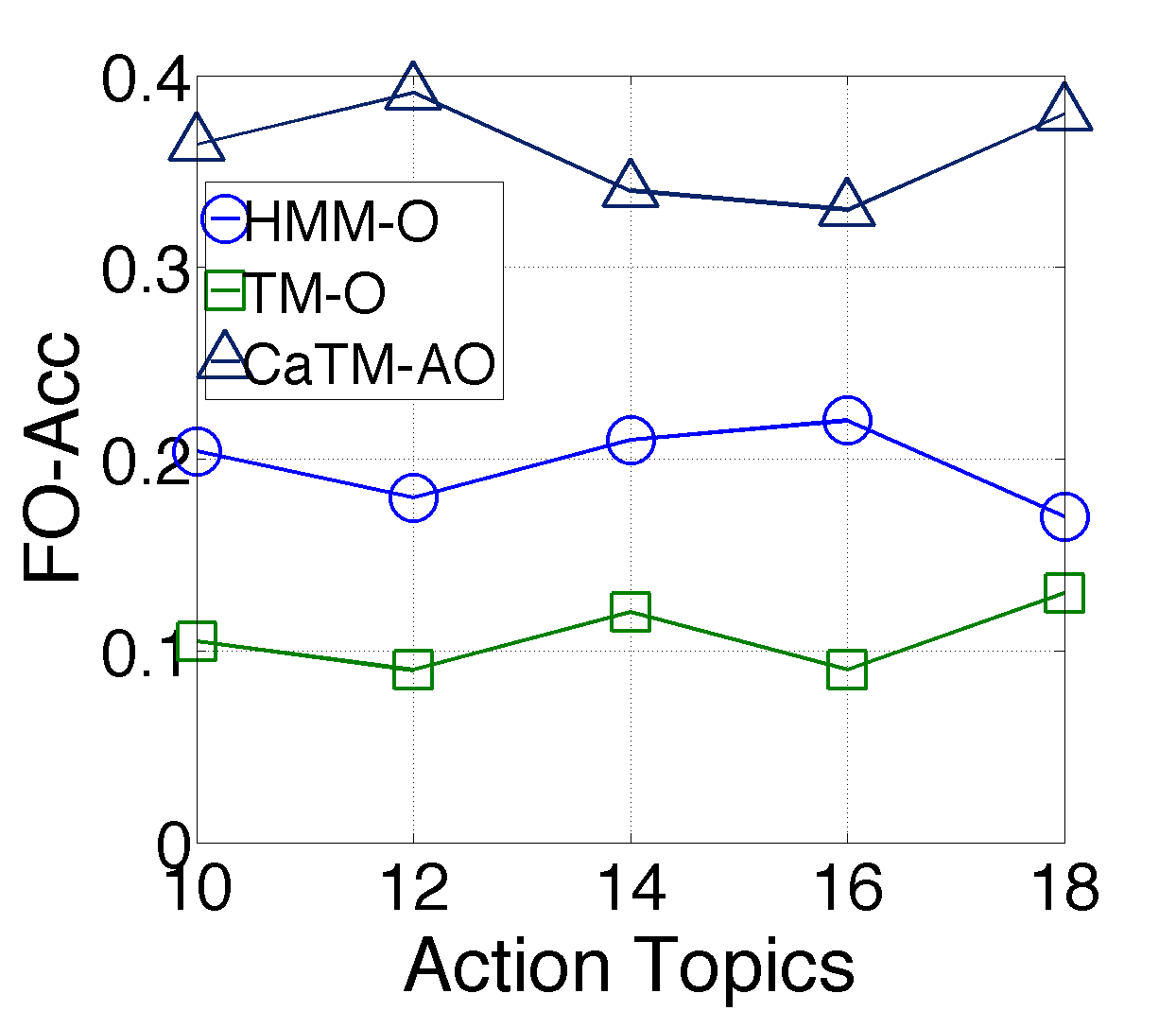}\hspace{-0.3in}
  \hspace{-0.1in}
 \caption{Forgotten action/object detection accuracy varied with the number of action-topics in the `office' dataset.} \label{fig:acp}
 \end{center}
\end{figure}

\begin{figure*}[t]
  \begin{center}
  \includegraphics[width=0.2\linewidth]{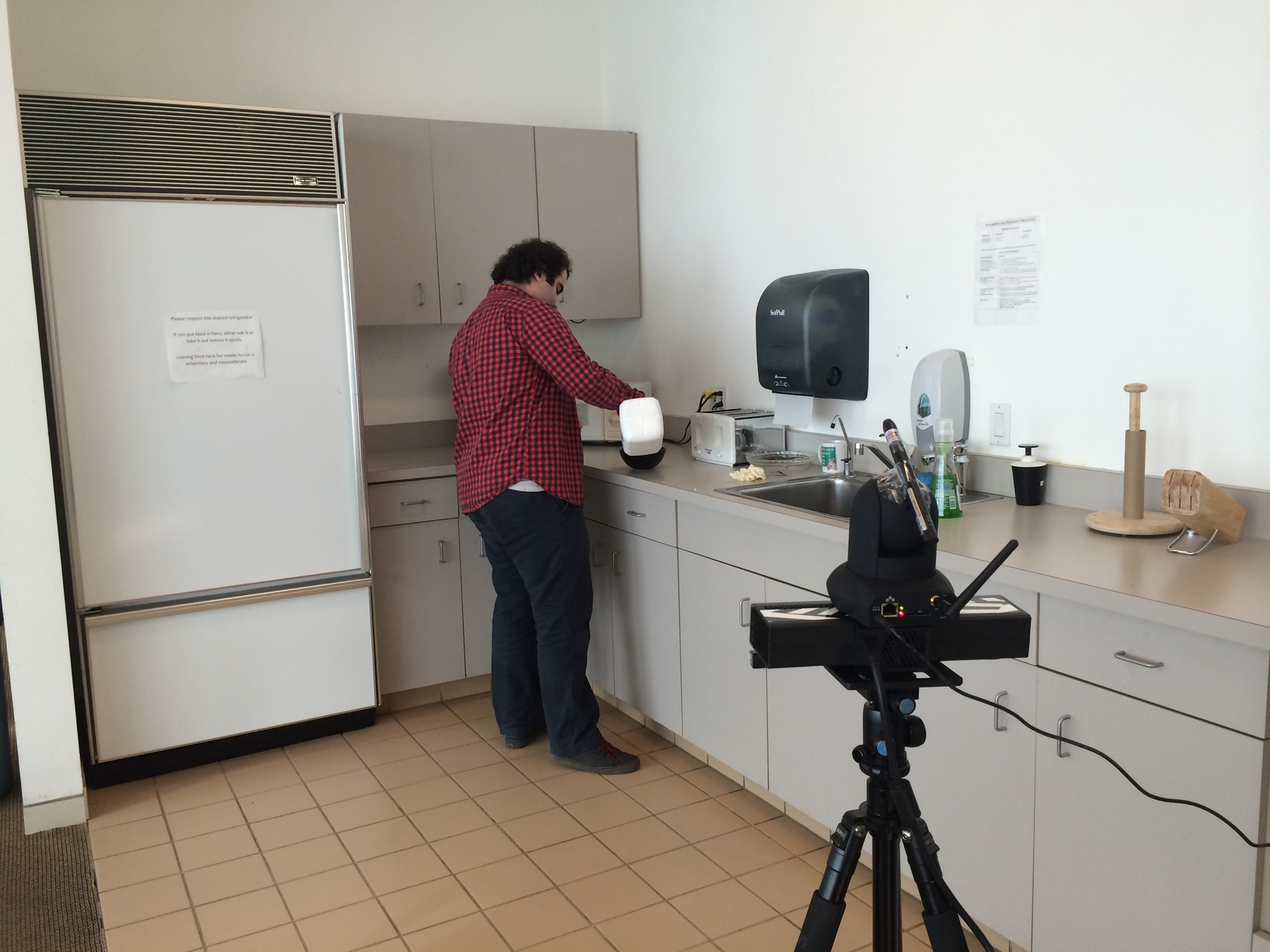}
  \includegraphics[width=0.2\linewidth]{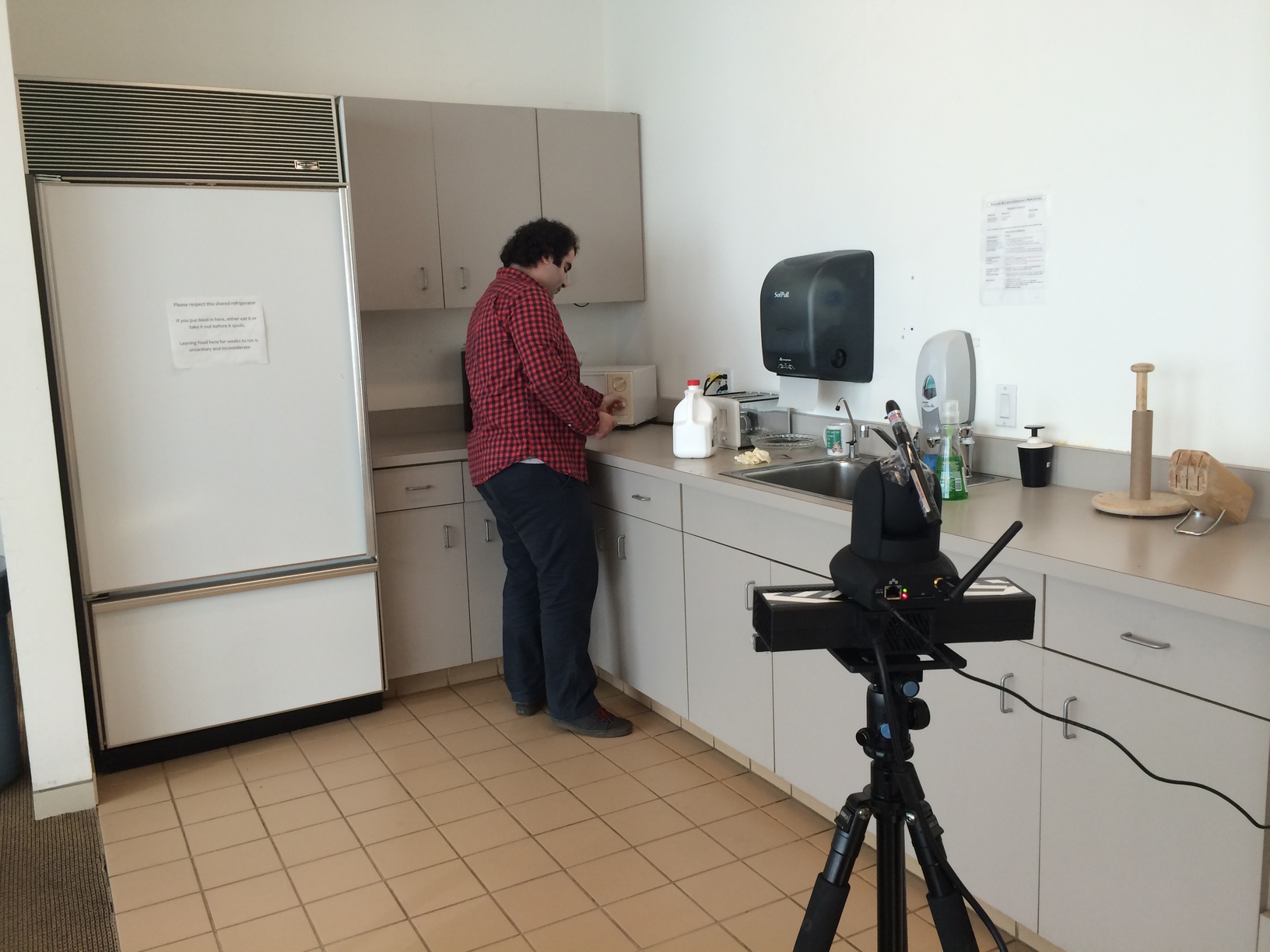}
  \includegraphics[width=0.2\linewidth]{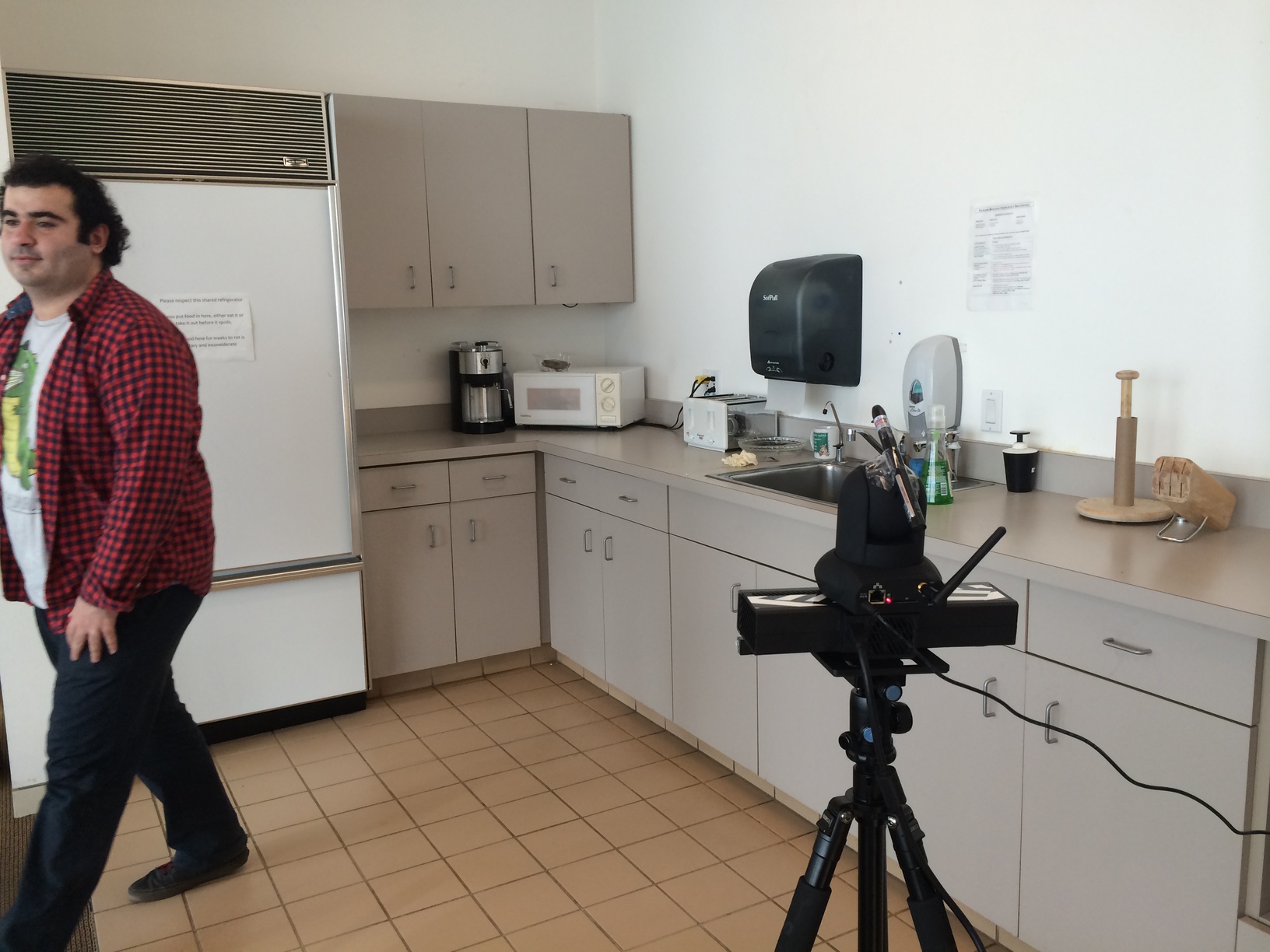}
  \includegraphics[width=0.2\linewidth]{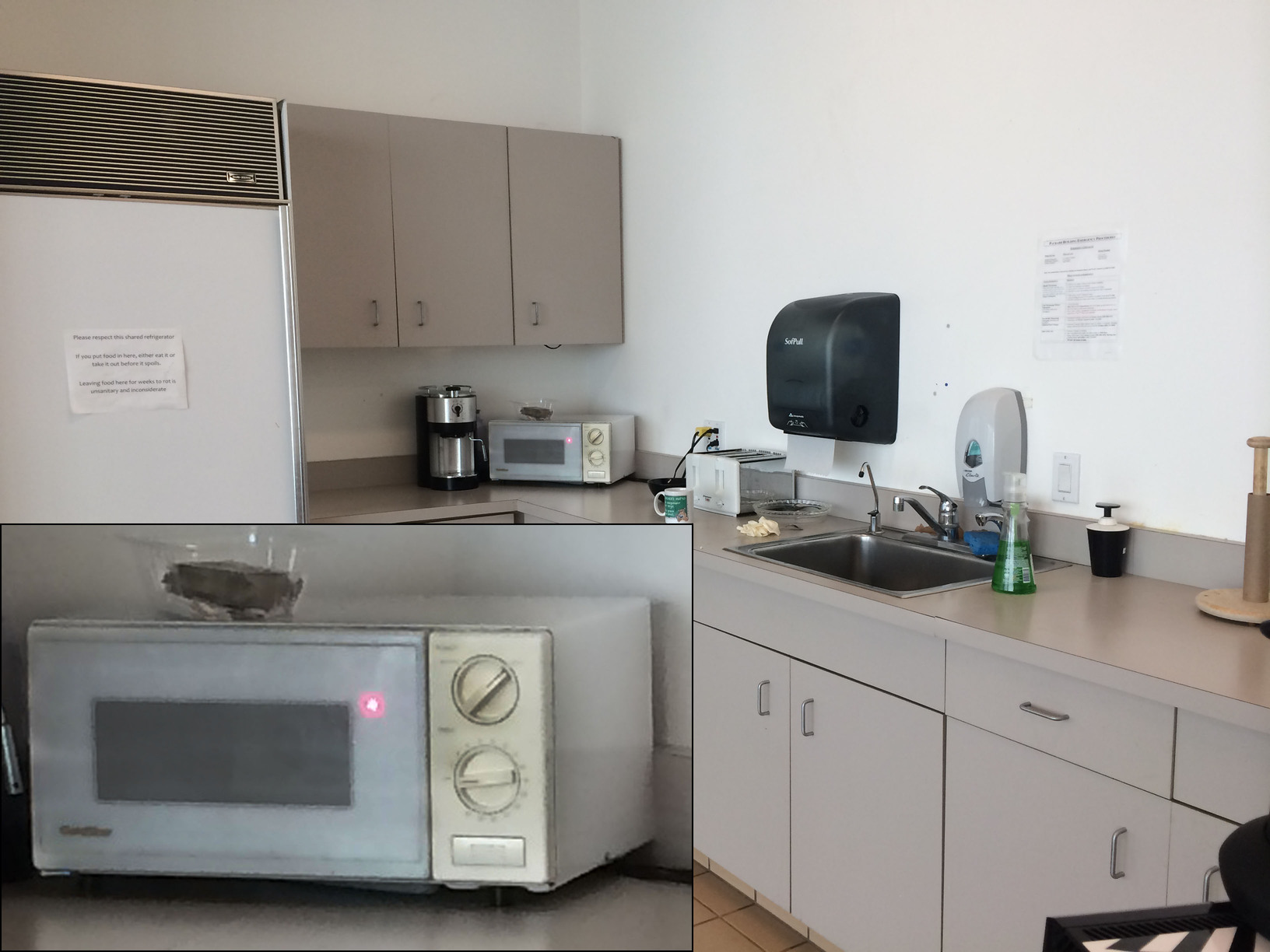}
  
 \caption{An example of the robotic experiment. The robot detects the human left the food in the microwave, then points to the microwave.} \label{fig:robotic}
 \end{center}
\end{figure*}

{\bf Did modeling the long-range relations help?}
We studied whether modeling the correlations and the temporal relations between topics was useful.
The approaches considering the temporal relations, HMM, TM-RT, and our CaTM, outperform other approaches which assume actions are temporal independent. This demonstrates that understanding temporal structure is critical to recognizing and patching actions. The approaches, TM-RT and CaTM, which model both the short-range and the long-range relations perform better than HMM only modeling local relations. Also, the approaches considering the topic correlations CTM, CTM-AT, and our CaTM perform better than the corresponding non-correlated topic models TM, TM-AT, and TM-RT. Our CaTM, which considers both the action correlation priors and the temporal relations, shows the best performance.


{\bf How successful was our unsupervised  approach in learning meaningful action-topics?} 
From Table~\ref{tb:re}, we can see that the unsupervised learned action-topics is promising to be semantically meaningful even though ground-truth semantic labels are not provided in the training.
In order to qualitatively estimate the performance, we give a visualization of our learned topics in Fig.~\ref{fig:vt}.  It shows that the actions with the same semantic meaning are clustered together though they are in different views and motions.
In addition to the one-to-one correspondence between topics and semantic action classes, we also plot the performance curves varied with the topic number in Fig.~\ref{fig:act}. It shows that if we set the topics a bit more than ground-truth classes, the performance increases since a certain action might be divided into multiple action-topics. But as topics increase, more variations are also introduced so that performance saturates.

{\bf RGB videos vs. \mbox{RGB-D} videos.}
In order to compare the effect of using information from \mbox{RGB-D} videos, we also evaluate
our model CaTM and HMM using the popular RGB features for action recognition (CaTM-A-DTF, CaTM-AO-DTF and HMM-DTF in Table~\ref{tb:re}). Clearly, the proposed human skeleton and \mbox{RGB-D} features outperform the DTF features as more accurate human motion and object are extracted.

{\bf How well did our new application of action patching performs?}
From Table~\ref{tb:re}, we find that the approaches learning the action relations mostly give better patching performance. This is because the learned co-occurrence and temporal structure strongly help indicate which actions are forgotten. Our model capturing both the short-range and long-range action relations shows the best results.

{\bf How important is it to consider relations between actions and objects?}
From the results, we can see that the model which did well in forgotten action detection also performed well in detecting forgotten object. Since our model CaTM-AO well considers the relations between the action and the object, it shows better performance in both forgotten action and forgotten object detection than those which models action and object independently as well as CaTM-A which only models the actions.


\begin{figure}[t]
  \begin{center}
  \includegraphics[width=1\linewidth]{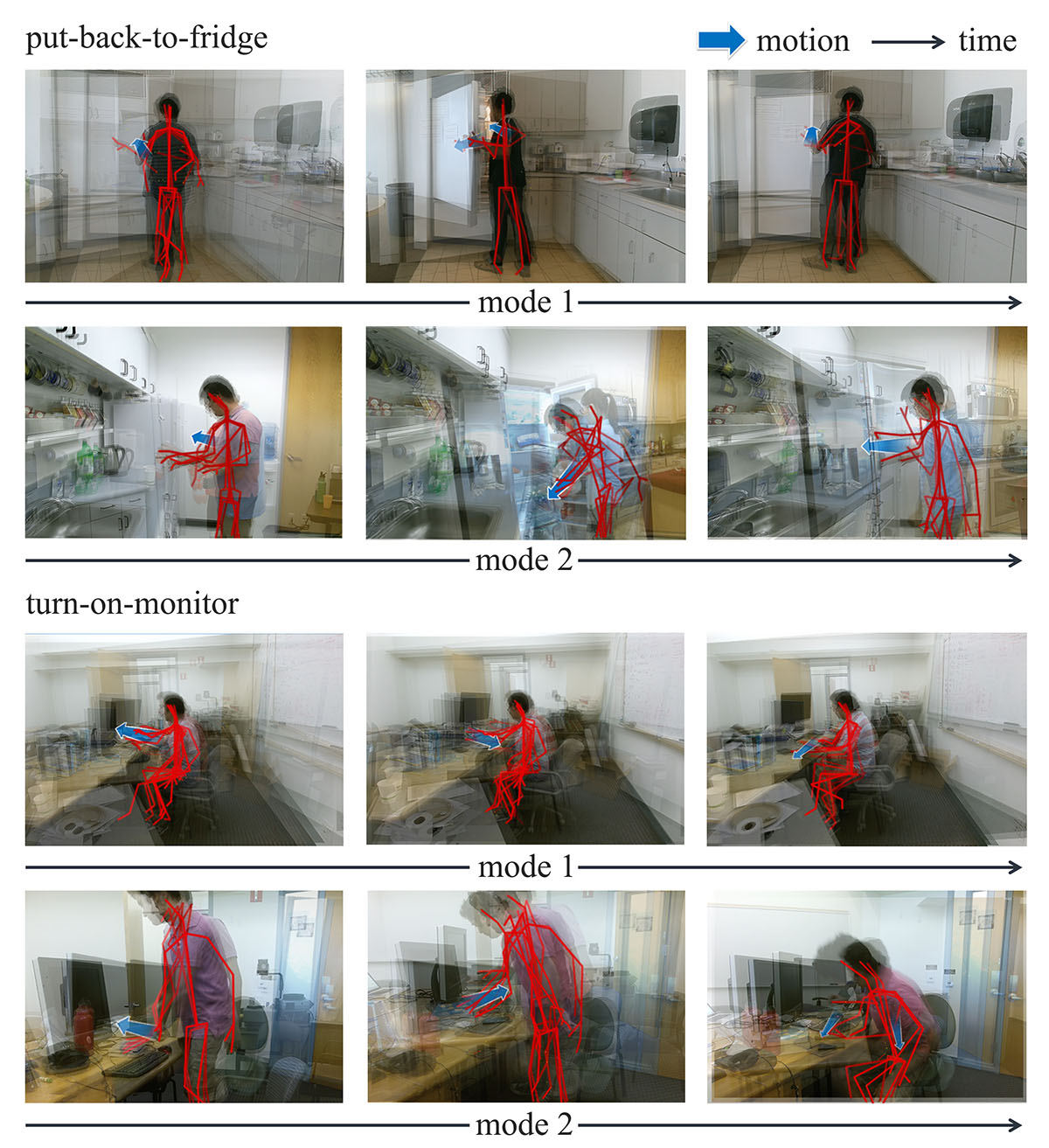}
  \caption{{\bf Visualization of the learned topics using our model.} For better illustration, we decompose the segments with the same topic into different modes (shown two) and divide a segment into three stages in time. The clips from different segments in the same stage are merged by scaling to the similar size of human skeletons.}
   \label{fig:vt}
 \end{center}
\end{figure}

\begin{table}[t]
\footnotesize
\setlength{\tabcolsep}{6pt}
\begin{center}
\caption{Robotic experiment results. The higher the better.}\label{tb:robore}
\begin{tabular*}{\linewidth}{@{\extracolsep{\fill}}c|c |c| c}
\hline
&Succ-Rate(\%)&Subj-AccScore(1-5)&Subj-HelpScore(1-5)\\
\hline\hline
HMM-O&37.5&2.1&2.3\\
TM-O&29.2&1.8&2.0\\
CaTM-AO&\textbf{62.5}&\textbf{3.5}&\textbf{3.9}\\
\hline
\end{tabular*}
\end{center}
\end{table}

\subsection{Robotic Experiments}
In this section, we show how our Watch-Bot reminds people of the forgotten actions in the real-world scenarios.  We test each two forgotten scenarios in `office' and `kitchen' respectively (\emph{put-back-book}, \emph{turn-off-monitor}, \emph{put-milk-back-to-fridge} and \emph{fetch-food-from-microwave}). We use a subset of the dataset to train the model in each activity type separately. In each scenario, we ask $3$ subjects to perform the activity twice. Therefore, we test $24$ trials in total. We evaluate three aspects. One is objective, the success rate (Succ-Rate): the laser spot lying within the object as correct. The other two are subjective, the average Subjective Accuracy Score (Subj-AccScore): we ask the participant if he thinks the pointed object is correct; and the average Subjective Helpfulness Score (Subj-HelpScore): we ask the participant if the output of the robot is helpful. Both of them are in $1-5$ scale, the higher the better.

Table~\ref{tb:robore} gives the results of our robotic experiments. We can see that our robot can achieve over $60\%$ success rate and gives the best performance. In most cases people think our robot is able to help them understand what is forgotten. Fig.~\ref{fig:robotic} gives an example of our experiment, in which our robot observed what a human is currently doing, realized he forgot to fetch food from microwave and then correctly pointed out the microwave in the scene.


\section{Conclusion and Future Work}\label{sec:con}

In this paper, we presented an algorithm that models the human activities in a completely unsupervised setting. We showed that it is important to modeling the long-range relations between the actions. To achieve this, we considered the video as a sequence of human-words/object-words, and an activity as a set of action-topics/object-topics. Then we modeled the word-topic distributions, the topic correlations and the topic relative time distributions. We then showed the effectiveness of our model in the unsupervised action segmentation and clustering, as well as the action patching. Moreover, we showed that our proposed robot system using the action patching algorithm was able to effectively remind people of forgotten actions in the real-world robotic experiments. For evaluation, we also contributed a new challenging \mbox{RGB-D} activity video dataset.

Though we showed the promising results and the interesting applications of the purely unsupervised models in the paper, we can see that the performance is not more than $50$ percent on the large-scale variant data, as we have no knowledge of the semantic information. In the future, we plan to extend the model to the semi-supervised approaches that can effectively use a small portion of the annotated data for better learning, and improve on the performance in the real-world applications.
\appendix


\begin{figure*}[t]
\begin{center}

  \subfigure[turn-on-monitor]{
  \begin{minipage}{0.32\linewidth}
  \includegraphics[width=1\linewidth]{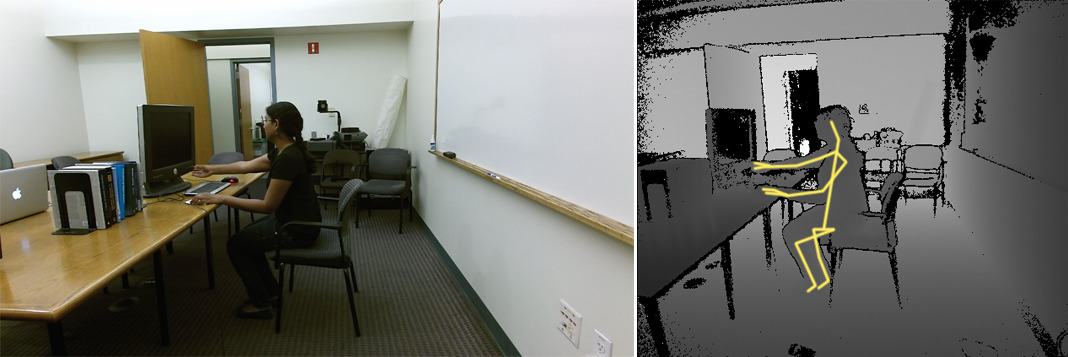}\vspace{0.01in}
  \end{minipage}
  }\hspace{-0.05in}
  \vspace{-0.05in}
  \subfigure[turn-off-monitor]{
  \begin{minipage}{0.32\linewidth}
  \includegraphics[width=1\linewidth]{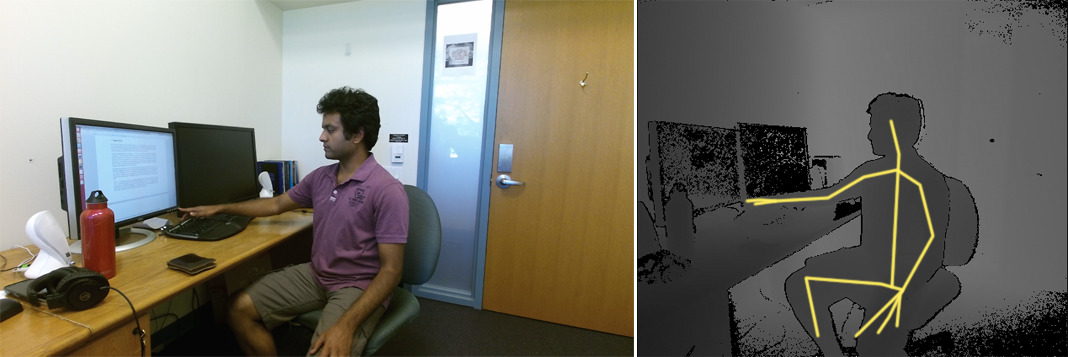}\vspace{0.01in} 
  \end{minipage}
  }\hspace{-0.05in}
  \subfigure[walk]{
  \begin{minipage}{0.32\linewidth}
  \includegraphics[width=1\linewidth]{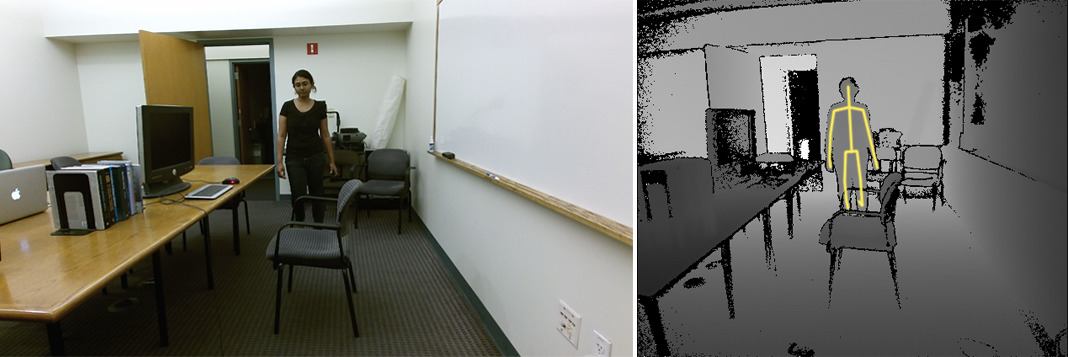}\vspace{0.01in}
  \end{minipage}
  }\hspace{-0.05in}
  
   \subfigure[play-computer]{
  \begin{minipage}{0.32\linewidth}
  \includegraphics[width=1\linewidth]{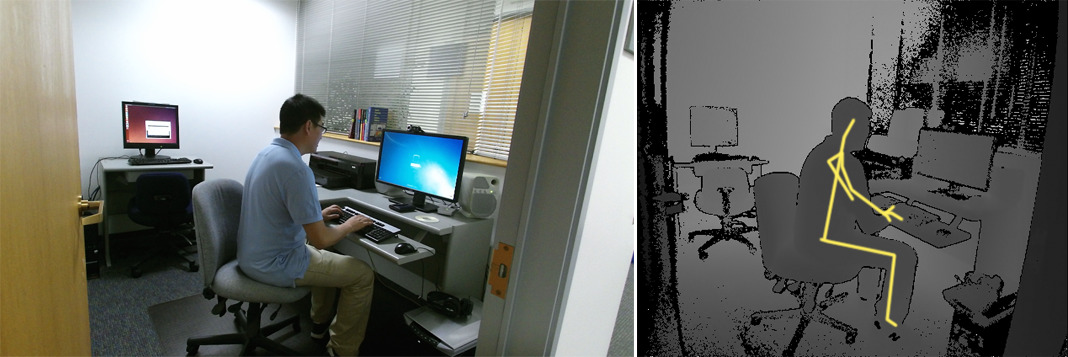}\vspace{0.01in}
  \end{minipage}
  }\hspace{-0.05in}
  \vspace{-0.05in}
  \subfigure[read]{
  \begin{minipage}{0.32\linewidth}
  \includegraphics[width=1\linewidth]{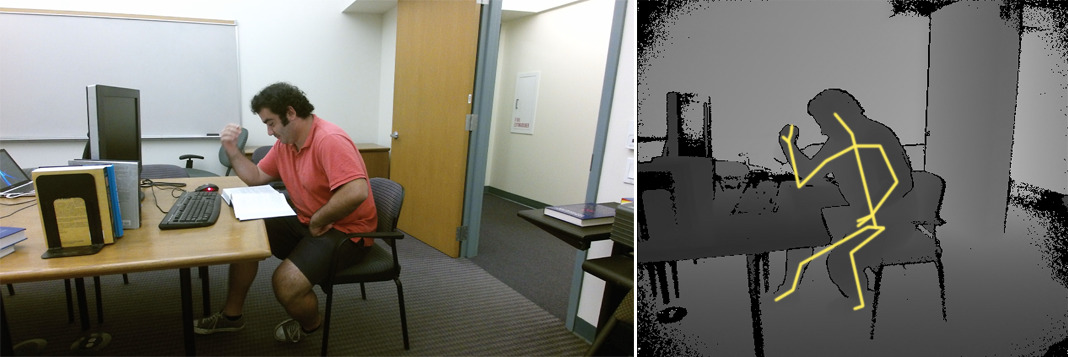}\vspace{0.01in} 
  \end{minipage}
  }\hspace{-0.05in}
  \subfigure[fetch-book]{
  \begin{minipage}{0.32\linewidth}
  \includegraphics[width=1\linewidth]{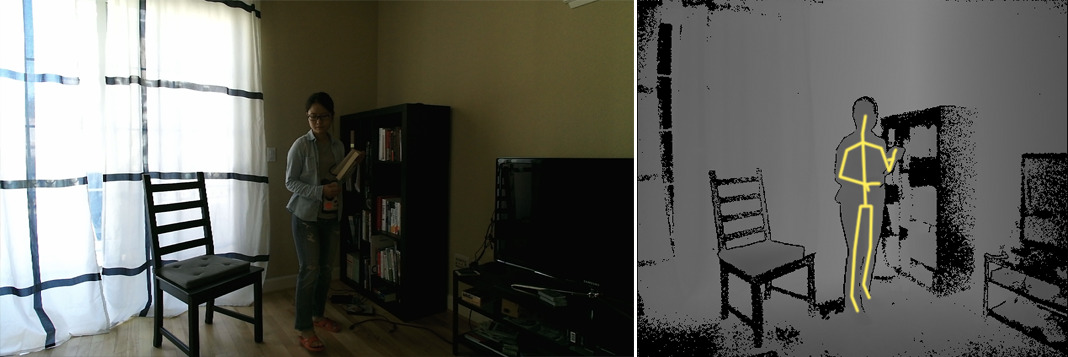}\vspace{0.01in}
  \end{minipage}
  }\hspace{-0.05in}
  
   \subfigure[put-back-book]{
  \begin{minipage}{0.32\linewidth}
  \includegraphics[width=1\linewidth]{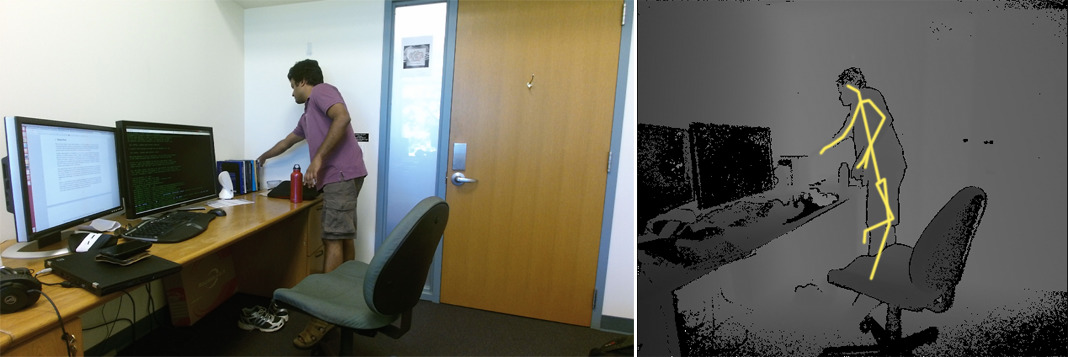}\vspace{0.01in}
  \end{minipage}
  }\hspace{-0.05in}
  \vspace{-0.05in}
  \subfigure[take-item]{
  \begin{minipage}{0.32\linewidth}
  \includegraphics[width=1\linewidth]{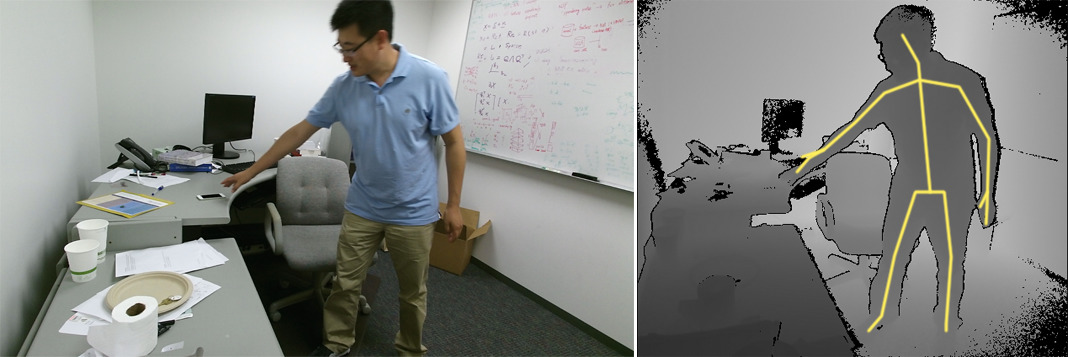}\vspace{0.01in} 
  \end{minipage}
  }\hspace{-0.05in}
  \subfigure[put-down-item]{
  \begin{minipage}{0.32\linewidth}
  \includegraphics[width=1\linewidth]{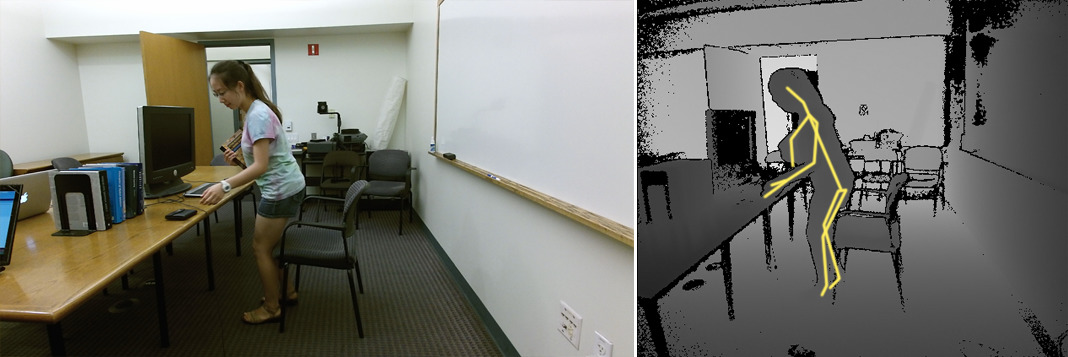}\vspace{0.01in}
  \end{minipage}
  }\hspace{-0.05in}
  
 \subfigure[leave-office]{
  \begin{minipage}{0.32\linewidth}
  \includegraphics[width=1\linewidth]{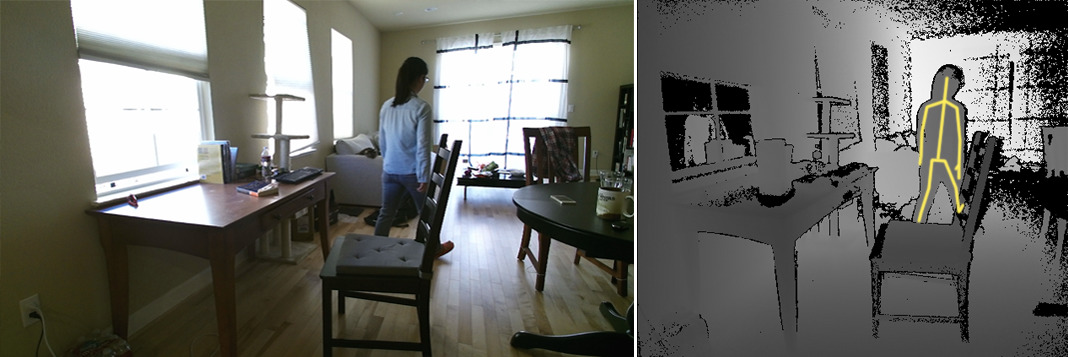}\vspace{0.01in}
  \end{minipage}
  }\hspace{-0.05in}
  \vspace{-0.05in}
  \subfigure[fetch-from-fridge]{
  \begin{minipage}{0.32\linewidth}
  \includegraphics[width=1\linewidth]{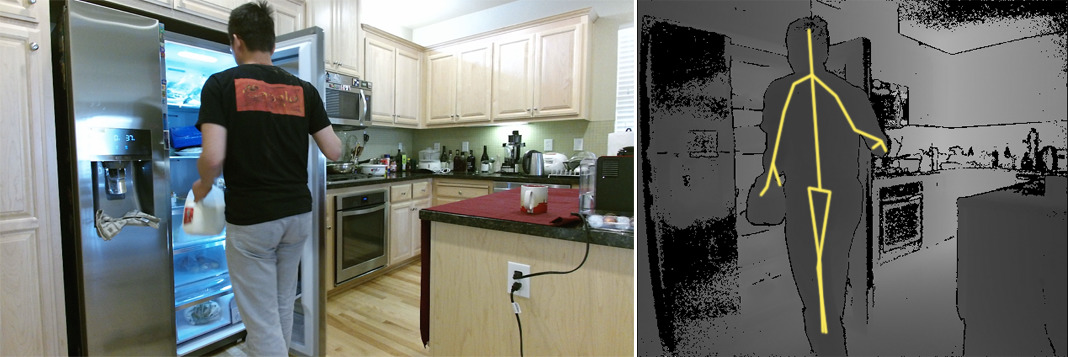}\vspace{0.01in} 
  \end{minipage}
  }\hspace{-0.05in}
  \subfigure[put-back-to-fridge]{
  \begin{minipage}{0.32\linewidth}
  \includegraphics[width=1\linewidth]{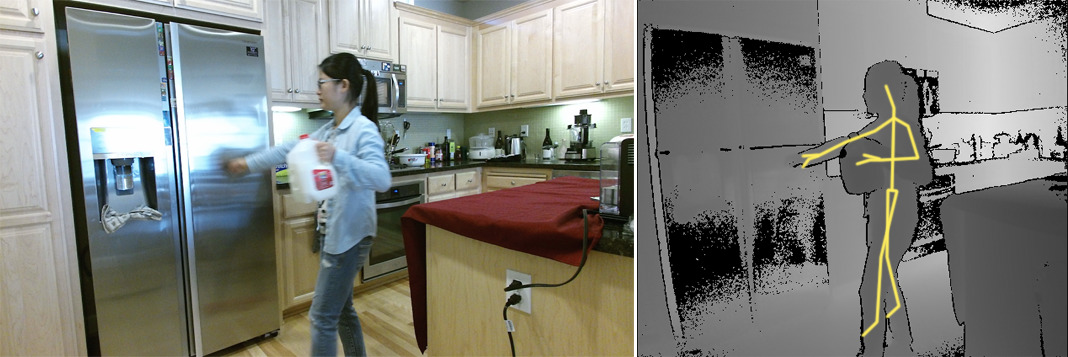}\vspace{0.01in}
  \end{minipage}
  }\hspace{-0.05in}
  
   \subfigure[prepare-food]{
  \begin{minipage}{0.32\linewidth}
  \includegraphics[width=1\linewidth]{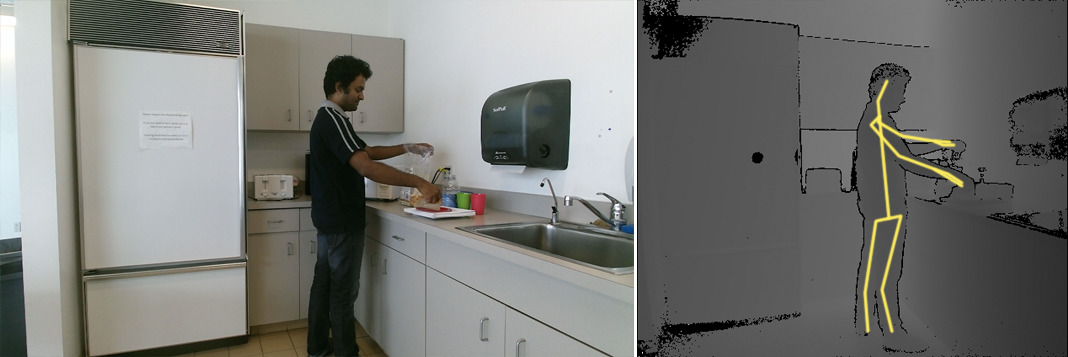}\vspace{0.01in}
  \end{minipage}
  }\hspace{-0.05in}
  \vspace{-0.05in}
  \subfigure[microwave]{
  \begin{minipage}{0.32\linewidth}
  \includegraphics[width=1\linewidth]{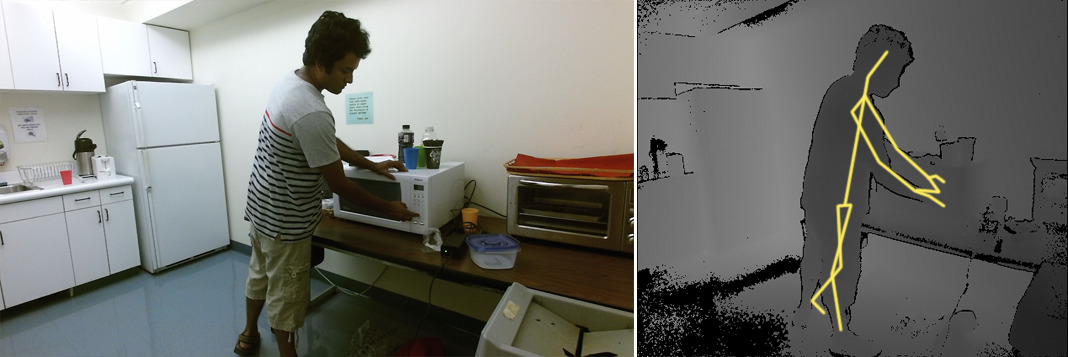}\vspace{0.01in} 
  \end{minipage}
  }\hspace{-0.05in}
  \subfigure[fetch-from-microwave]{
  \begin{minipage}{0.32\linewidth}
  \includegraphics[width=1\linewidth]{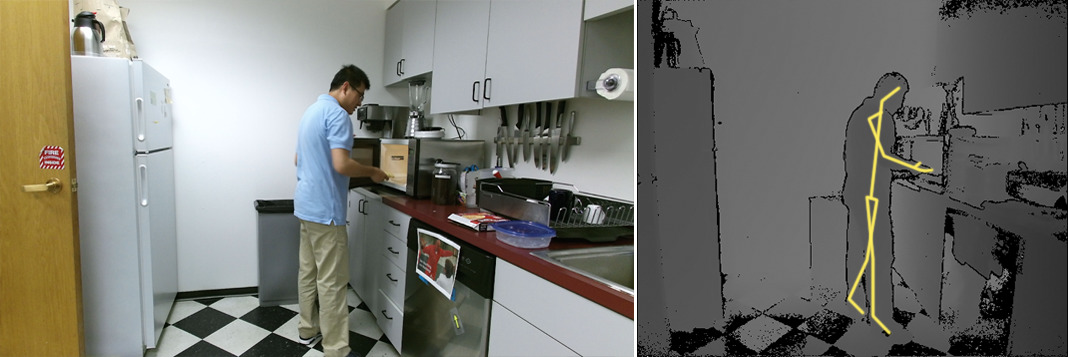}\vspace{0.01in}
  \end{minipage}
  }\hspace{-0.05in}

   \subfigure[pour]{
  \begin{minipage}{0.32\linewidth}
  \includegraphics[width=1\linewidth]{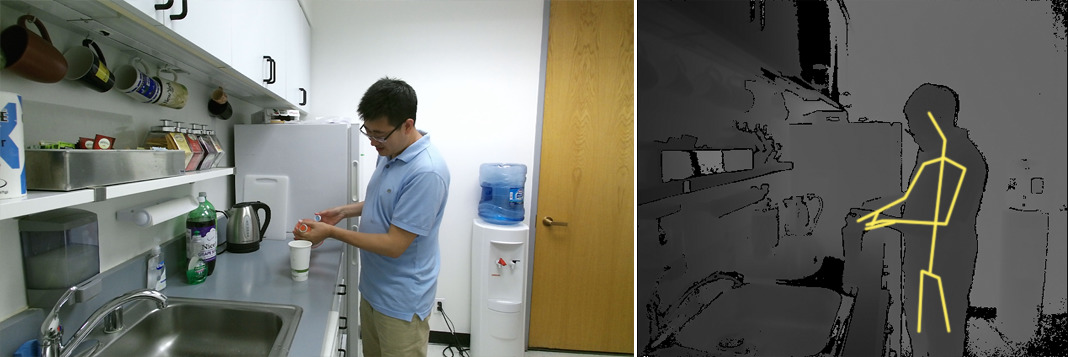}\vspace{0.01in}
  \end{minipage}
  }\hspace{-0.05in}
  \vspace{-0.05in}
  \subfigure[drink]{
  \begin{minipage}{0.32\linewidth}
  \includegraphics[width=1\linewidth]{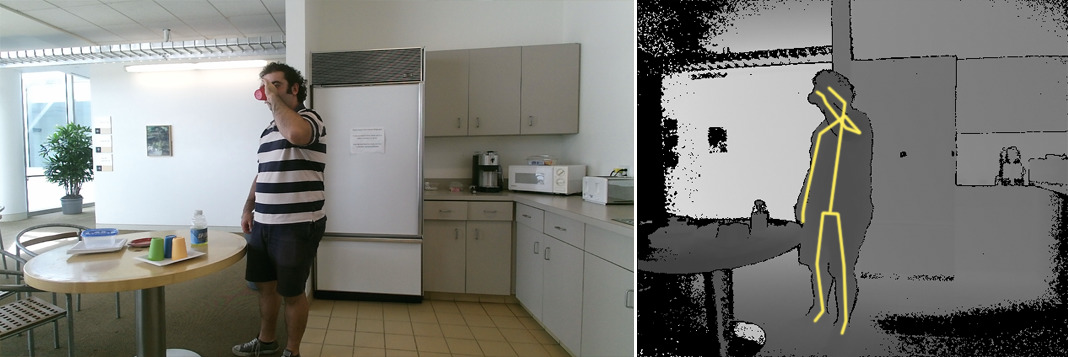}\vspace{0.01in} 
  \end{minipage}
  }\hspace{-0.05in}
  \subfigure[leave-kitchen]{
  \begin{minipage}{0.32\linewidth}
  \includegraphics[width=1\linewidth]{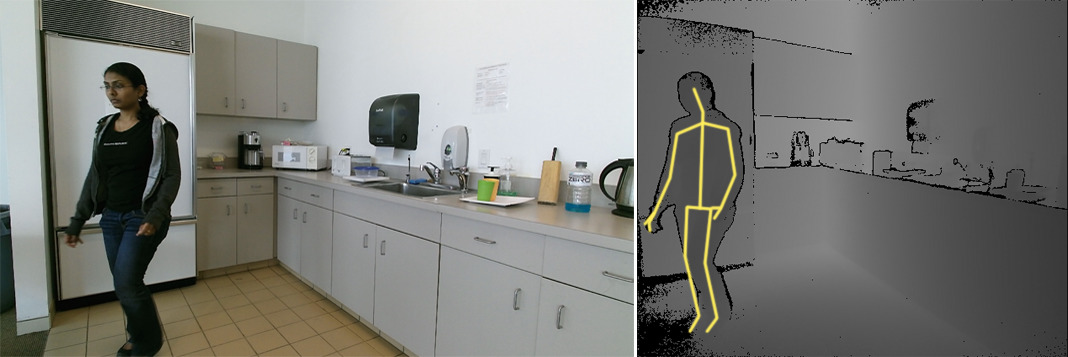}\vspace{0.01in}
  \end{minipage}
  }\hspace{-0.05in}
  
   \subfigure[move-kettle]{
  \begin{minipage}{0.32\linewidth}
  \includegraphics[width=1\linewidth]{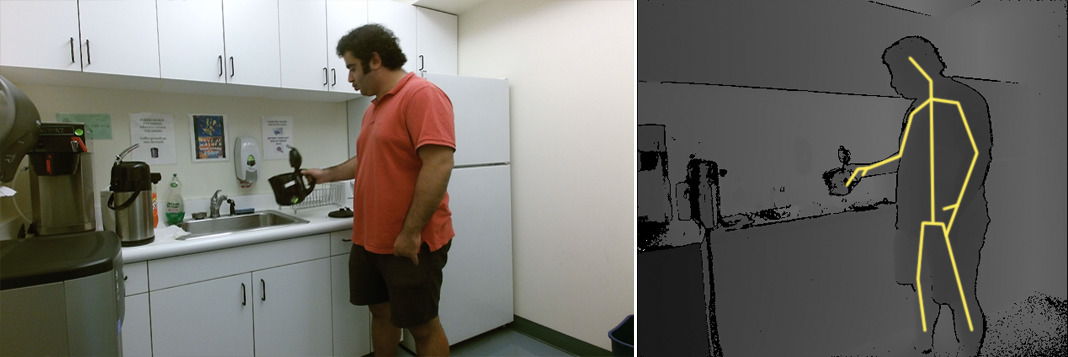}\vspace{0.01in}
  \end{minipage}
  }\hspace{-0.05in}
  \vspace{-0.05in}
  \subfigure[fill-kettle]{
  \begin{minipage}{0.32\linewidth}
  \includegraphics[width=1\linewidth]{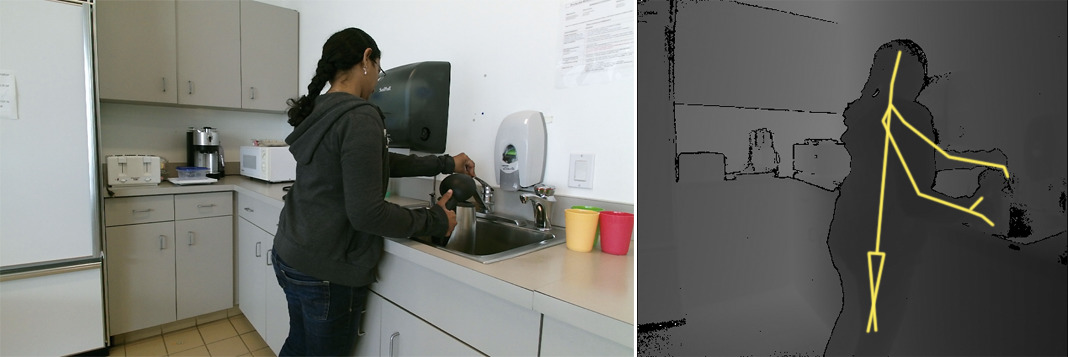}\vspace{0.01in} 
  \end{minipage}
  }\hspace{-0.05in}
  \subfigure[plug-in-kettle]{
  \begin{minipage}{0.32\linewidth}
  \includegraphics[width=1\linewidth]{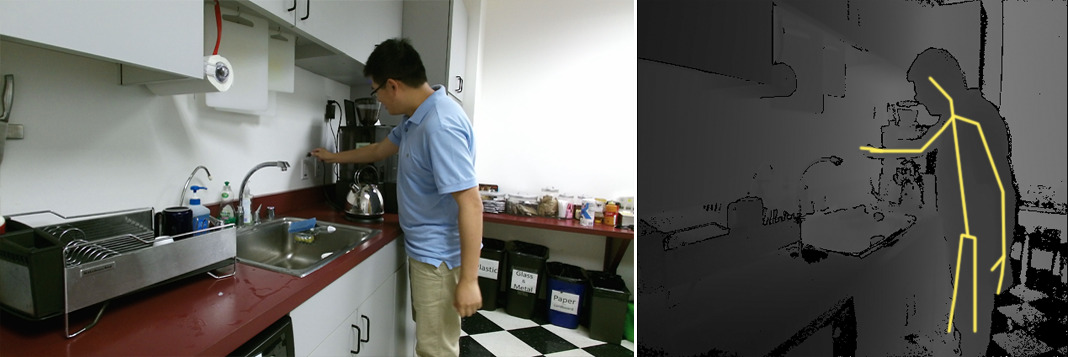}\vspace{0.01in}
  \end{minipage}
  }\hspace{-0.05in}

  \caption{Examples of every action class in our dataset. The left is RGB frame and the right is depth frame with human skeleton (yellow).}
   \label{fig:actions}
 \end{center}
\vspace*{-0.1in}
\end{figure*}

\section{\\Derivation of Gibbs Sampling}\label{app:dev}
We give the detailed derivation of the posterior distribution of $z_{nd}$ (Eq.~(\ref{eqn:posz})) in this section. We begin with the joint distribution $p(\w^h,\w^o,\mathbf{t},\z^{(1)},\z^{(2)}|\pi^{(1)},\pi^{(2)},\beta^{(1)},\beta^{(12)},\theta)$, where $\w^h,\w^o,\mathbf{t},\z^{(1)},\z^{(2)},\pi^{(1)},\pi^{(2)}$ are all variables of the word $w^{h}_{nd}$,$w^{o}_{nd}$, the time stamp of a word $t_{nd}$, the topic-assignment of a word $z^{(1)}_{nd}$,$z^{(2)}_{nd}$ and the topic probability $\pi^{(1)}_{kd}$,$\pi^{(2)}_{kd}$ in $D$ documents of $K$ action topics and $P$ object topics.
\vspace{0.05in}
\begin{align*}
&p(\w^h,\w^o,\mathbf{t},\z^{(1)},\z^{(2)}|\pi^{(1)},\pi^{(2)},\beta^{(1)},\beta^{(12)},\theta)\\
=\ &p(\w^h|\z^{(1)},\beta^{(1)})p(\w^o|\z^{(1)},\z^{(2)},\beta^{(12)})\cdot\\
&p(\mathbf{t}|\z^{(1)},\theta)p(\z^{(1)}|\pi^{(1)})p(\z^{(2)}|\pi^{(2)})\\
=\ &\int p(\w^h|\z^{(1)},\phi^{(1)})p(\phi^{(1)},\beta^{(1)})d\phi^{(1)} \cdot \\
&\int p(\w^o|\z^{(1)},\z^{(2)},\phi^{(12)})p(\phi^{(12)},\beta^{(12)})d\phi^{(12)} \cdot \\
&p(\mathbf{t}|\z^{(1)},\theta) \cdot p(\z^{(1)}|\pi^{(1)})p(\z^{(2)}|\pi^{(2)}).
\end{align*}
where the joint distribution is decided by the following five terms.

topic-word distributions:
\begin{align*}
&\int p(\w^h|\z^{(1)},\phi^{(1)})p(\phi^{(1)},\beta^{(1)})d\phi^{(1)}\\
=&\int\prod_{d=1}^D\prod_{n=1}^{N_d}\phi^{(1)}_{z^{(1)}_{nd},w^h_{nd}}\prod_{k=1}^K\frac{1}{B(\beta^{(1)})}\prod_w\phi_{kw}^{(1)\beta^{(1)}_w-1}d\phi^{(1)}_k\\
=&\prod_{k=1}^K\frac{1}{B(\beta^{(1)})}\int\prod_w\phi_{kw}^{(1)N_{kw}+\beta^{(1)}_w-1}d\phi^{(1)}_k\\
=&\prod_{k=1}^K\frac{B(N_{k}+\beta^{(1)})}{B(\beta^{(1)})}\\
&\int p(\w^o|\z^{(1)},\z^{(2)},\phi^{(12)})p(\phi^{(12)},\beta^{(12)})d\phi^{(12)}\\
=&\prod_{k=1}^K\prod_{p=1}^P\frac{B(N_{kp}+\beta^{(12)})}{B(\beta^{(12)})},
\end{align*}
where we denote the Beta function as $B(\beta)=\frac{\prod_{k=1}^K\Gamma(\beta_k)}{\Gamma(\sum_{k=1}^K\beta_k)}$. 

topic-pair relative time distribution:
 \begin{align*}
&p(\mathbf{t}|\z^{(1)},\theta)=\prod_{d=1}^D\prod_{n=1}^{N_d}p(t_{nd}|z^{(1)}_{:d},\theta)\\
&\quad \quad \quad \ \ =\prod_{d=1}^D\prod_{m=1}^{N_d}\prod_{n=1}^{N_d}p(t_{mnd}|\theta_{z^{(1)}_{md},z^{(1)}_{nd}}).\\
\end{align*}

topic priors:
\begin{align*}
&p(\z^{(1)}|\pi^{(1)})=\prod_{d=1}^D\prod_{n=1}^{N_d}\pi^{(1)}_{z^{(1)}_{nd},d}\\
&p(\z^{(2)}|\pi^{(2)})=\prod_{d=1}^D\prod_{n=1}^{N_d}\pi^{(2)}_{z^{(2)}_{nd},d}.
\end{align*}

Then for a certain assignment $z^{(1)}_{nd}$, we give the posterior using the above joint distribution:
\vspace{0.05in}
\begin{align*}
&p(z^{(1)}_{nd}|\pi^{(1)}_{:d},z^{(1)}_{-nd},z^{(2)}_{nd},t_{nd})\\
=&\frac{p(\w^h,\w^o,\mathbf{t},\z^{(1)},\z^{(2)}|\pi^{(1)},\pi^{(2)},\beta^{(1)},\beta^{(12)},\theta)}{p(\w^h,\w^o,\mathbf{t},z^{(1)}_{-nd},\z^{(2)}|\pi^{(1)},\pi^{(2)},\beta^{(1)},\beta^{(12)},\theta)}\\
\propto &\frac{p(\w^h,\w^o,\mathbf{t},\z^{(1)},\z^{(2)}|\pi^{(1)},\pi^{(2)},\beta^{(1)},\beta^{(12)},\theta)}{p(w^h_{-nd},w^o_{-nd},t_{-nd},z^{(1)}_{-nd},\z^{(2)}|\pi^{(1)},\pi^{(2)},\beta^{(1)},\beta^{(12)},\theta)}\\
=&\pi^{(1)}_{z^{(1)}_{nd},d}\omega(z^{(1)}_{nd},w^h_{nd})\omega(z^{(1)}_{nd},z^{(2)}_{nd},w^o_{nd})p(t_{nd}|z^{(1)}_{:d},\theta),
\end{align*}
\vspace{0.05in}
where:

\begin{align*}
&\omega(z^{(1)}_{nd},w^h_{nd})=\prod_{k=1}^K\frac{B(N_{k}+\beta^{(1)})}{B(N_{k}^{-nd}+\beta^{(1)})}=\frac{N_{z^{(1)}_{nd},w^h}^{-nd}+\beta^{(1)}}{N_{z^{(1)}_{nd}}^{-nd}+N_{w^h}\beta^{(1)}}\\
&\omega(z^{(1)}_{nd},z^{(2)}_{nd},w^o_{nd})=\prod_{k=1}^K\prod_{p=1}^P\frac{B(N_{kp}+\beta^{(12)})}{B(N_{kp}^{-nd}+\beta^{(12)})}\\
&\quad \quad \quad \quad \quad \quad \quad \  \ =\frac{N_{z^{(1)}_{nd},z^{(2)}_{nd},w^o}^{-nd}+\beta^{(12)}}{N_{z^{(1)}_{nd},z^{(2)}_{nd}}^{-nd}+N_{w^o}\beta^{(12)}}\\
&p(t_{nd}|z^{(1)}_{:d},\theta)=\prod_m p(t_{mnd}|\theta_{z^{(1)}_{md},z^{(1)}_{nd}})p(t_{nmd}|\theta_{z^{(1)}_{nd},z^{(1)}_{md}})\\ 
&\quad \quad \quad \quad \quad \  \ =\prod_m \Omega(t_{mnd}|\theta_{z^{(1)}_{md},z^{(1)}_{nd}})\Omega(t_{nmd}|\theta_{z^{(1)}_{nd},z^{(1)}_{md}}).
\end{align*}
\vspace{0.05in}
Then assign $z^{(1)}_{nd}$ with a specific topic $k$, we have the sampling posterior Eq.~(\ref{eqn:posz}):
\vspace{0.05in}
\begin{align*}
&p(z^{(1)}_{nd}=k|\pi^{(1)}_{:d},z^{(1)}_{-nd},z^{(2)}_{nd},t_{nd})\notag\\
&\propto \pi^{(1)}_{kd}\omega(k,w^h_{nd})\omega(k,z^{(2)}_{nd},w^o_{nd})p(t_{nd}|z^{(1)}_{:d},\theta),\notag\\
&\omega(k,w^h_{nd})=\frac{N_{kw^h}^{-nd}+\beta^{(1)}}{N_{k}^{-nd}+N_{w^h}\beta^{(1)}},\notag\\
&\omega(k,p,w^o_{nd})=\frac{N_{kpw^o}^{-nd}+\beta^{(12)}}{N_{kp}^{-nd}+N_{w^o}\beta^{(12)}},\notag\\
&p(t_{nd}|z^{(1)}_{:d},\theta)=\prod_m^{N_d} \Omega(t_{mnd}|\theta_{z^{(1)}_{md},k})\Omega(t_{nmd}|\theta_{k,z^{(1)}_{md}}),
\end{align*}
\vspace{0.05in}
Similarly we have:
\begin{align*}
&p(z^{(2)}_{nd}=p|\pi^{(2)}_{:d},z^{(2)}_{-nd},z^{(1)}_{nd})\propto
\pi^{(2)}_{pd}\omega(z^{(1)}_{nd},p,w^o_{nd}).
\end{align*}
\vspace{0.05in}
\begin{table*}[t]
\setlength{\tabcolsep}{14pt}
\centering
\caption{Action sequence examples in our dataset. The action `walk' is omitted as it can be between any actions. Possible forgotten actions are in the brackets. }\label{tb:data}
\begin{tabular}{ll}
\hline
office:\\\hline
read $\rightarrow$ leave-office\\
fetch-book $\rightarrow$ read $\rightarrow$ [put-back-book] $\rightarrow$ leave-office\\
put-down-item $\rightarrow$ read $\rightarrow$ [take-item] $\rightarrow$ leave-office\\
put-down-item $\rightarrow$ fetch-book $\rightarrow$ read $\rightarrow$ [put-back-book] $\rightarrow$ [take-item] $\rightarrow$ leave-office\\
put-down-item $\rightarrow$ fetch-book $\rightarrow$ read $\rightarrow$ [take-item] $\rightarrow$ [put-back-book] $\rightarrow$ leave-office\\
fetch-book $\rightarrow$ put-down-item $\rightarrow$ read $\rightarrow$ [put-back-book] $\rightarrow$ [take-item] $\rightarrow$ leave-office\\
fetch-book $\rightarrow$ put-down-item $\rightarrow$ read $\rightarrow$ [take-item] $\rightarrow$ [put-back-book] $\rightarrow$ leave-office\\
turn-on-monitor $\rightarrow$ play-computer $\rightarrow$ [turn-off-monitor] $\rightarrow$ leave-office\\
put-down-item $\rightarrow$ turn-on-monitor $\rightarrow$ play-computer $\rightarrow$ [turn-off-monitor] $\rightarrow$ [take-item] $\rightarrow$ leave-office\\
put-down-item $\rightarrow$ turn-on-monitor $\rightarrow$ play-computer $\rightarrow$ [take-item] $\rightarrow$ [turn-off-monitor] $\rightarrow$ leave-office\\
turn-on-monitor $\rightarrow$ put-down-item $\rightarrow$ play-computer $\rightarrow$ [turn-off-monitor] $\rightarrow$ [take-item] $\rightarrow$ leave-office\\
turn-on-monitor $\rightarrow$ put-down-item $\rightarrow$ play-computer $\rightarrow$ [take-item] $\rightarrow$ [turn-off-monitor] $\rightarrow$ leave-office\\
\hline
kitchen:\\\hline
pour $\rightarrow$ drink\\ 
pour $\rightarrow$ [drink] $\rightarrow$ leave-kitchen\\
fetch-from-fridge $\rightarrow$ pour $\rightarrow$ put-back-to-fridge $\rightarrow$ drink \\
fetch-from-fridge $\rightarrow$ pour $\rightarrow$ [put-back-to-fridge] $\rightarrow$ [drink] $\rightarrow$ leave-kitchen\\
fetch-from-fridge $\rightarrow$ pour $\rightarrow$ drink $\rightarrow$ put-back-to-fridge\\
fetch-from-fridge $\rightarrow$ pour $\rightarrow$ [drink] $\rightarrow$ [put-back-to-fridge] $\rightarrow$ leave-kitchen\\
fetch-from-fridge $\rightarrow$ prepare-food $\rightarrow$ [put-back-to-fridge] $\rightarrow$ microwave $\rightarrow$ [fetch-from-microwave] $\rightarrow$ leave-kitchen\\
fetch-from-fridge $\rightarrow$ prepare-food $\rightarrow$ microwave $\rightarrow$ [put-back-to-fridge] $\rightarrow$ [fetch-from-microwave] $\rightarrow$ leave-kitchen\\
fetch-from-fridge $\rightarrow$ prepare-food $\rightarrow$ microwave $\rightarrow$ [fetch-from-microwave] $\rightarrow$ [put-back-to-fridge] $\rightarrow$ leave-kitchen\\
move-kettle $\rightarrow$ fill-kettle $\rightarrow$ move-kettle  $\rightarrow$ [plug-in-kettle] \\
\hline
\end{tabular}
\end{table*}
\bibliographystyle{ieee}
\bibliography{wpatch}

\end{document}